\documentclass{amsart}
\usepackage{setspace}
\usepackage[margin=1in]{geometry}
\usepackage{graphicx}
\usepackage{bmpsize} 
\usepackage{bm}
\usepackage{multirow}
\usepackage{mathtools}
\usepackage[shortlabels]{enumitem}
\usepackage{amssymb}
\usepackage[ruled,vlined]{algorithm2e}

\usepackage[T1]{fontenc}
\usepackage[utf8]{inputenc}
\usepackage[english]{babel}
\usepackage{graphbox}

\newtheorem{theorem}{Theorem}[section]

\theoremstyle{definition}

\theoremstyle{definition}
\newtheorem{remark}[theorem]{Remark}

\newcommand{\RN}[1]{%
	\textup{\uppercase\expandafter{\romannumeral#1}}%
}
\newcommand{\bb}[1]{\mathbf{#1}}
\newcommand{\IP}[3]{\left\langle #2, #3 \right\rangle_{#1}}
\newcommand{\nn}[1]{\left\|#1\right\|}
\newcommand{\vp}{\varphi}
\newcommand{\sig}{\bm{\sigma}}
\newcommand{\mmu}{\bm{\mu}}
\newcommand{\pphi}{\bm{\varphi}}
\newcommand{\ppsi}{\bm{\psi}}

\newcommand{\rev}[1]{{#1}}

\makeatletter
\newcommand{\addresseshere}{%
  \enddoc@text\let\enddoc@text\relax
}
\makeatother

\title{A Comparison of Neural Network Architectures for Data-Driven Reduced-Order Modeling}
\author{Anthony Gruber$^{1,*}$}
\author{Max Gunzburger$^1$}
\author{Lili Ju$^2$}
\author{Zhu Wang$^2$}
\thanks{$^*$Corresponding author: (Anthony Gruber)  agruber@fsu.edu}

\email{agruber@fsu.edu, mgunzburger@fsu.edu, ju@math.sc.edu, wangzhu@math.sc.edu}

\address{$^1$ Department of Scientific Computing, Florida State University, 400 Dirac Science Library, Tallahassee, FL 32306, USA}

\address{$^2$ Department of Mathematics, University of South Carolina, 1523 Greene Street, Columbia, SC 29208, USA}

\begin{document}

\maketitle

\begin{abstract}
    The popularity of deep convolutional autoencoders (CAEs) has engendered \rev{new and} effective reduced-order models (ROMs) for the simulation of large-scale dynamical systems.  \rev{Despite this}, it is still unknown whether deep CAEs provide superior performance \rev{over established linear techniques or other network-based methods} in all modeling scenarios.  To elucidate this, the effect of autoencoder architecture on its associated ROM is studied through the comparison of deep CAEs against two alternatives: a simple fully connected autoencoder, and a novel graph convolutional autoencoder.  Through benchmark experiments, it is shown that the superior autoencoder architecture for a given ROM application is highly dependent on the size of the latent space and the structure of the snapshot data, with the proposed architecture demonstrating benefits on data with irregular connectivity when the latent space is sufficiently large.
    
    \vspace{0.5pc}


\end{abstract}


\section{Introduction}
High-fidelity computational models are indispensable tools for the prediction and analysis of physical systems which are governed by parameterized partial differential equations (PDEs).  As more and more industries are relying on simulated results to inform their daily operations, the significant amount of computational resources demanded by such models is becoming increasingly prohibitive.  Indeed, actions which increase model fidelity such as refining the spatio-temporal resolution can also lead to an explosion of dimensionality, making use of the full-order model (FOM) infeasible in real-time or many-query scenarios.  To remedy this, emphasis has been placed on reduced-order models (ROMs) which approximate the high-fidelity, full-order models at any desired configuration of parameters.  Indeed, an appropriately built ROM can enable important applications such as uncertainty quantification or predictive control, which respectively require access to the FOM solution at thousands of points or near-instantaneous access to approximate solutions.

The governing assumption inherent in the design of ROMs is that the FOM solution lies in a submanifold of the ambient space which is intrinsically low-dimensional.  This is reasonable to posit when the FOM solution has been generated from the semi-discretization of an evolving quantity over a dense set of nodal points, as the values of this quantity in each dimension are strongly determined by the original (presumably lower-dimensional) dynamics.  When this is the case, efficient approximation of the solution submanifold can lead to massive decreases in simulation time with relatively small approximation errors, since restriction to this space is lossless in principle.  On the other hand, the problem of computing this submanifold is generally nontrivial, which has led to a wide variety of ROMs appearing in practice. 

To enable this speedup in approximation of the FOM solution, ROMs are divided into offline and online stages.  The offline stage is where the majority of expensive computation takes place, and may include operations such as collecting snapshots of the high-fidelity model, generating a reduced basis, or training a neural network.  On the other hand, an effective ROM has an online stage that is designed to be as fast as possible, involving only a small amount of function evaluations or the solution of small systems of equations.  Classically, most ROMs have relied on some notion of reduced basis to accomplish this task, which is designed to capture all of the important information about the high-fidelity system.  For example, the well studied method of proper orthogonal decomposition (POD) described in Section~\ref{sec:prelims} uses a number of solution snapshots to generate a reduced basis of size $n$ whose span has minimal $\ell_2$ reconstruction error.  The advantage of this is that the high-fidelity problem can then be projected into the reduced space, so that only its low-dimensional representation must be manipulated at solving time.  On the other hand, POD like many reduced-basis methods is a fundamentally linear technique, and therefore struggles in the presence of advection-dominated phenomena where the solution \rev{decays slowly} over time.  

\subsection{Related Work}

Desire to break the accuracy barrier of linear methods along with the rapid advancement of neural network technology has reignited interest in machine learning methods for model order reduction.  In particular, it was first noticed in Milano and Koumoutsakos \cite{milano2002} that the reduced basis built with a linear multi-layer perceptron (MLP) model is equivalent to POD, and that nonlinear MLP offers improved prediction and reconstruction abilities at additional computational cost.  Later, this idea was extended in works such as \cite{kashima2016,hartman2017} which established that a fully connected autoencoder architecture can be trained to provide a nonlinear analogue of projection which often realizes significant performance gains over linear methods such as POD \rev{when the reduced dimension is small.}

Following the mainstream success of large-scale parameter sharing network architectures such as convolutional neural networks, the focus of nonlinear ROMs began to shift in this direction.  Since many high-fidelity PDE systems such as large-scale fluid simulations involve a huge number of degrees of freedom, the ability to train an effective network at a fraction of the memory cost for a fully connected network is very appealing and led to several important breakthroughs.  The work of Lee and Carlberg in \cite{lee2020} demonstrated that deep convolutional autoencoders (CAEs) can overcome the Kolmogorov width barrier for advection-dominated systems which limits linear ROM methods, leading to a variety of similar ROMs based on this architecture seen in e.g. \cite{fukami2020,eivazi2020,fresca2021, maulik2021}.  Moreover, works such as \cite{eivazi2020,fresca2021} have experimented with entirely data-driven ROMs based on deep CAEs, and seen success using either fully connected or recurrent long short term networks to simulate the reduced low-dimensional dynamics.  At present, there are a large variety of nonlinear ROMs based on fully connected and convolutional autoencoder networks, and this remains an active area of research.

Despite the now widespread use of autoencoder networks for reduced-order modeling, there have not yet been studies which compare the performances of fully connected and convolutional networks in a ROM setting.  In fact, many high-fidelity simulations take place on domains which are unstructured and irregular, so it is somewhat surprising that CAEs still perform well for this type of problem (see e.g. \cite[Remark 5]{fresca2021}).  On the other hand, there is clearly a conceptual issue with applying the standard convolutional neural network (c.f. Section~\ref{sec:prelims}) to input data with variable connectivity; since the convolutional layers in this architecture accept only rectangular data, the input must be reshaped in a way which can change its neighborhood relationships, meaning the convolution operation may no longer be localized in space.  This is a known issue that has been thoroughly considered in the literature on graph convolutional neural networks (see e.g. \cite{wu2020} and references therein), which have been successful in a wide range of classification tasks.

\subsection{Contributions}
The goal of this work is to augment the ROM literature with a comparison between ROM-autoencoder architectures based on fully connected and convolutional neural networks.  In particular, we compare the most popular deep convolutional autoencoder architecture from e.g. \cite{lee2020,fresca2021} to two alternatives:  a basic fully connected autoencoder with batch normalization \cite{ioffe2015}, and a novel autoencoder based on the GCNII graph convolution operation of \cite{chen2020}.  Through benchmark experiments on regular and irregular data, we study the effectiveness of each architecture in compressing/decompressing high-fidelity PDE data, and evaluate its performance as part of a data-driven ROM.  Results show that the superior architecture is highly dependent on the size of the reduced latent space as well as the modeling task involved.  Moreover, some of the assumed advantages of the CAE (such as reduced memory consumption and easier training) are not satisfied in every case, and for some applications it is advantageous to use one of the other available architectures \rev{or a linear technique such as POD.}  It is our hope that the experiments provided here will enable modelers to make a more informed choice regarding which network architecture they employ for their autoencoder-based ROM.






\section{Preliminaries}\label{sec:prelims}
Consider a full-order model which is a dynamical system of parameterized ODEs,
\begin{equation}\label{eq:fom}
    \dot{\bb{x}}(t,\mmu) = \bb{f}\left(t,\bb{x}(t,\mmu),\mmu\right), \qquad \bb{x}(0,\mmu) = \bb{x}_0(\mmu).
\end{equation}
Here $t\in[0,T]$ is time ($T>0$), $\mmu\in D\subset \mathbb{R}^{n_\mu}$ is the vector of model parameters, $\bb{x}: [0,T]\times D \to \mathbb{R}^N$ is the parameterized state function with initial value $\bb{x}_0\in\mathbb{R}^N$, and $\bb{f} \in [0,T] \times \mathbb{R}^N \times D \to \mathbb{R}^N$ is the state velocity.  In practice, models \eqref{eq:fom} arise naturally from the semi-discretization of a parameterized system of PDE, where the discretization of the domain creates an $N$-length vector (or array) $\bb{x}$ representing a quantity defined at the nodal points.  Clearly, a semi-discretization of size $N$ has the potential to create a large amount of redundant dimensionality, as a unique dimension is created for every nodal point while the solution remains characterized by the same number of parameters.

\begin{figure}
    \centering
    \includegraphics[width=0.5\linewidth]{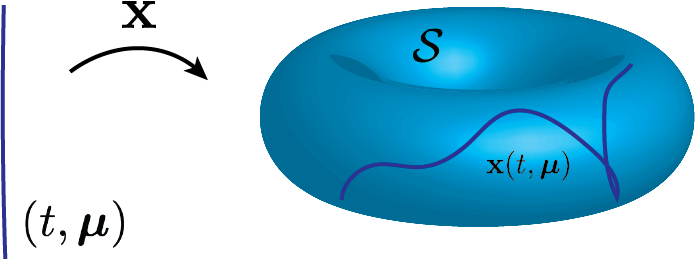}
    \caption{An illustration of the solution submanifold $\mathcal{S}\subset\mathbb{R}^n$ parametrized by the quantities $(t,\mmu)$.}
    \label{fig:lowdim}
\end{figure}

To effectively handle the high-dimensional ODE \eqref{eq:fom}, the overarching goal of reduced-order models is to predict approximate solutions without solving the governing equations in the high-dimensional space $\mathbb{R}^N$.  In principle, this is possible as long as the assignment $(t,\mmu) \mapsto \bb{x}(t,\mmu)$ is unique for each $(t,\mmu)$, meaning that the space of solutions spans a low-dimensional manifold in the ambient $\mathbb{R}^N$ (c.f. Figure~\ref{fig:lowdim}).  Therefore, the ROM should provide fast access to the solution manifold
\[ \mathcal{S} = \{\bb{x}(t,\mmu)\,\,|\,\,t\in[0,T],\,\mmu\in D \} \subset \mathbb{R}^N,\]
which contains integral curves of the high-dimensional FOM.  Note that the uniqueness assumption on $\bb{x}$ implies that the coordinates $(t,\mmu)$ form a global chart for $\mathcal{S}$ and hence $\mathcal{S}$ is intrinsically $(n_\mu + 1)$-dimensional (at most), \rev{which is frequently much less than the ambient dimension $N$.}

Traditionally, many ROMs are generated through some form of Proper Orthogonal Decomposition (POD).  To apply POD, some number of solutions to \eqref{eq:fom} are first collected and stored in a snapshot matrix $\bb{S} = (s^i_j) \in \mathbb{R}^{N\times N_s}$ where $N_s$ is the number of training samples, $s^i_j = x^i(t_j,\mmu_j)$, $x^i$ represents the $i^{th}$ component of the solution $\bb{x}$, and nonuniqueness is allowed in the parameters $\mmu_j$.  A singular value decomposition of the snapshot matrix $\bb{S}=\bb{U}\bm{\Sigma}\bb{V}^\top$ is then performed, and the first $n$ columns $\bb{U}_n$ of $\bb{U}\in\mathbb{R}^{N\times N}$ (corresponding to the largest $n$ eigenvalues of $\Sigma$) are chosen to serve as a reduced basis for the solution space.  Geometrically, this procedure amounts to choosing as basis an $n$-plane of POD modes in $\mathbb{R}^N$ (the columns of $\bb{U}_n$) which minimizes the $\ell_2$ projection error $\left\|\bb{x} - \bb{U}_n\bb{U}_n^\top\bb{x}\right\|$ over the snapshot data.  A ROM can then be generated from this by projecting \eqref{eq:fom} to this $n$-plane.  More precisely, writing the approximate solution $\tilde{\bb{x}} = \bb{U}_n\hat{\bb{x}}$ for some $\hat{\bb{x}}\in\mathbb{R}^n$ and using the FOM along with the fact that $\bb{U}_n^\top\bb{U}_n = \bb{I}_n$ leads to the system
\[ \dot{\hat{\bb{x}}}(t,\mmu) = \bb{U}_n^\top\bb{f}\left(t,\bb{U}_n\hat{\bb{x}},\mmu\right), \qquad \hat{\bb{x}}(0,\mmu) = \hat{\bb{x}}_0(\mmu) = \bb{U}_n^\top\bb{x}_0(\mmu), \]
which requires the solution of only $n\ll N$ equations.

POD based techniques are known to be highly effective in many cases, especially when the problem dynamics are diffusion-dominated.  On the other hand, advection-dominated problems exhibit a slowly decaying Kolmogorov n-width
\[ d_n(\mathcal{S}) = \inf_{M_n}\sup_{\bb{x}\in\mathcal{S}}\inf_{\hat{\bb{x}}\in M_n} \|\bb{x}-\hat{\bb{x}}\|,\]
meaning a large number of POD modes is required for accurate approximation of their solutions.  This reveals a fundamental issue with linear approximation spaces which motivated recent study into neural network based ROMs.  By now, it is well known that the barrier posed by $d_n$ can be surmounted using deep convolutional autoencoders, which are now reviewed.


\subsection{Convolutional Autoencoder Networks}
Recall that a fully connected neural network (FCNN) is a function $\bb{f} = \bb{f}_n \circ \bb{f}_{n-1} \circ ... \circ \bb{f}_1$ of the input vector $\bb{x} = \bb{x}_0 \in \mathbb{R}^{N_0}$ such that at layer $\ell$,
\begin{equation}\label{eq:cnn}
    \bb{x}_\ell \coloneqq \bb{f}_\ell(\bb{x}_{\ell-1}) =  \sig_\ell\left(\bb{W}_\ell\bb{x}_{\ell-1} + \bb{b}_\ell\right),
\end{equation}
where $\sig_\ell = \mathrm{Diag}(\sigma_\ell)$ is a nonlinear activation function acting row-wise on its input, $\bb{W}_\ell \in \mathbb{R}^{{N_\ell} \times N_{\ell-1}}$ is the weight matrix of the layer and $\bb{b} \in \mathbb{R}^{N_\ell}$ is its bias.  FCNNs are extremely expressive due to their large amount of parameters, but for the same reason they are typically difficult to train and incur a high memory cost.  On the other hand, a convolutional neural network (CNN) can be considered a special case of FCNN where weights are shared and determined by a translation-invariant kernel function.  More precisely, a convolutional neural network is a function of a multidimensional input $\bb{x}=\bb{x}_0$ such that at layer $\ell$,
\[ \bb{x}_{\ell,i} = \sig_\ell\left(\sum_{j=1}^{C_{in}}\bb{x}_{(\ell-1),j} \ast \bb{W}^j_{\ell,i} + \bb{b}_{\ell,i} \right), \]
where $1\leq i \leq C_{out}$ denotes the spatially-independent ``channel dimension'' of $\bb{x}$ \rev{which may differ at each layer (i.e. $C_{in}\neq C_{out})$},  $\bb{W}_{\ell,i}^j$ is the convolution kernel at layer $\ell$, and $\bb{b}_{\ell,i}$ is a bias object which is constant for each fixed $\ell,i$. Note that the kernel $\bb{W}$ has tensor dimension equal to that of the input plus one, since it depends on the shape of $\bb{x}_{\ell-1}$ as well as the number of channels at layers $\ell, \ell-1$.  Here $\ast$ plays the role of the discrete convolution operator, defined for two-dimensional, one-channel arrays $\bb{x},\bb{W}$ as 
\[ \left(\bb{x} \ast \bb{W}\right)^\alpha_{\beta} = \sum_{\gamma,\delta} x^{(s\alpha+\gamma)}_{(s\beta+\delta)}\, w^{(L-1-\gamma)}_{(M-1-\delta)}, \]
where $s \in \mathbb{N}$ is the stride length,  $\bb{W}=(w_\delta^\gamma)\in \mathbb{R}^L \times \mathbb{R}^M$, and all Greek indices are compactly supported and begin at zero.  It follows from this definition of the discrete convolution that the ranges of $\alpha, \beta$ in the output depend on the constants $s, M, L$ as well as the shape of the input.  To determine this more precisely, consider projecting the tensor $\bb{x}\ast\bb{W}$ along some coordinate direction and computing the length of this projection in terms of the lengths of the corresponding projections of $\bb{x}, \bb{W}$.  If the projection of $\bb{x}$ along this direction has length $I$ and the projection of $\bb{W}$ has length $F\leq I$, then it is straightforward to show \cite[Relationship 5]{dumoulin2016} that 
the output projection has length $1 + \left\lfloor\frac{I-F}{s}\right\rfloor$.  Therefore, convolution will naturally decrease the dimensionality of the input data $\bb{x}$ when $F\geq 2$, creating a downsampling effect which is highly useful for techniques such as image compression. This downsampling can be further controlled by padding the input $\bb{x}$ with an array of zeros, in which case the size of the output changes based on other calculable rules (see e.g. \cite{dumoulin2016}).

Since the discrete convolution \eqref{eq:cnn} decreases the dimensionality of the input data in a controlled way, it is reasonable to make use of it when constructing convolutional autoencoder (CAE) networks.   In particular, the main idea behind CAE networks is to stack convolutional layers in such a way that the input $\bb{x}$ is successively downsampled (or encoded) to a low-dimensional representation, and then upsampled (or decoded) back to its original form.  Combined with an appropriate notion of reconstruction error to be minimized, this encourages the neural network to learn an approximation to the identity mapping $\bb{x} \approx \tilde{\bb{x}} = \bb{g}\circ \bb{h}(\bb{x})$ which is useful for tasks such as compressed sensing and PDE approximation provided the decoder $\bb{g}$ can be decoupled from the encoder $\bb{h}$.  In the context of ROM, the notion of data downsampling serves as a nonlinear form of projection analogous to the truncated eigenvalue decomposition in POD.  In principle, as more characteristic features are extracted by the convolutional layers in the network, less dimensionality is required to encode the original solution $\bb{x}$.  Therefore, repeated convolution enables the most basic features of $\bb{x}$ to be represented with a latent space of small dimension, which can be exploited by the network for accurate reconstruction of the solution manifold.  Moreover, the assumption of uniqueness in the solution mapping $(t,\mmu)\mapsto \bb{x}(t,\mmu)$ guarantees that this idea is  loss-less whenever the latent dimension is at least as large as $n_\mu+1$.  

On the other hand, if convolution is to be useful for ROM, it is necessary to have a way to upsample (or decode) the latent representation for $\bb{x}$.  Although the convolution operation is clearly not invertible, it can be transposed in a way which is useful for this purpose.  In particular, note that discrete convolution can be considered a matrix operation by lexicographically ``unrolling'' the inputs $\bb{x}$ into a vector and similarly unrolling the kernel $\bb{W}$ into a sparse matrix which acts on that vector.  The operator $\bb{W}$ then has a well defined transpose, which effectively broadcasts the (unrolled) low-dimensional input to a high-dimensional output which can be rolled back into its original shape.  In practice, this operation is accomplished by simply exchanging the forward and backward passes of the relevant convolutional layer, as it can be shown (see \cite[Section 4]{dumoulin2016}) that this is operationally equivalent to computing the transposed convolution.  Doing this several times in succession produces the required upsampling effect, and traditionally makes up the decoder part of the CAE structure.

\subsection{Autoencoder-based ROMs}

As mentioned before, using a well trained CAE $\bb{g}\circ\bb{h}$ in place of the linear reconstruction $\bb{U}_n\bb{U}_n^\top$ when designing a ROM can provide a great benefit to accuracy.  In particular, examples from \cite{lee2020,fresca2021} show that deep CAEs can approach the theoretical minimum error with only a few latent variables, enabling improved accuracy over POD-ROMs \rev{when the number of modes is small.}  Moreover, autoencoder-based ROMs do not require any more information than standard POD-ROMs.  Indeed, developing a ROM using the decoder mapping $\bb{g}$ means seeking $\tilde{\bb{x}}\approx \bb{x}$ which satisfies 
\[ \tilde{\bb{x}}(t,\mmu) = \bb{g}\left(\hat{\bb{x}}(t, \mmu)\right), \]
where $\hat{\bb{x}}:\mathbb{R}^{n_\mu + 1} \to \mathbb{R}^n$, $n \ll N$ is a latent representation of the state $\bb{x}$,  and $\bb{g}:\mathbb{R}^n \to \mathbb{R}^N$ is a nonlinear mapping from the latent space to the ambient $\mathbb{R}^N$.  In particular, if $\bb{g}$ can be constructed such that its image is the solution submanifold $\mathcal{S}$, then the comparatively small latent representation $\hat{\bb{x}}$ is sufficient for complete characterization of the state $\bb{x}$.  As $\bb{g}$ is typically trained using only the snapshots in $\mathbb{S}$, when $n\geq n_\mu+1$  this provides the potential for an accurate nonlinear mapping using the same information necessary for POD.


Using the approximation $\tilde{\bb{x}}$ in the FOM \eqref{eq:fom}, it follows that $\dot{\tilde{\bb{x}}} = \bb{g}'(\hat{\bb{x}})\dot{\hat{\bb{x}}}$ where $\bb{g}'(\hat{\bb{x}})$ denotes the derivative operator (i.e. Jacobian matrix) of $\bb{g}$.  Therefore, a reduced order model involving only $\hat{\bb{x}},\bb{g}$ can be generated by assuming that the approximate state $\tilde{\bb{x}}$ obeys the same dynamics as the original state $\bb{x}$.  In particular, if $\dot{\tilde{\bb{x}}} = \bb{f}$ and $\bb{g}'$ has full rank at every $\hat{\bb{x}}$, it follows quickly that
\begin{equation}\label{eq:rom}
    \dot{\hat{\bb{x}}}(t,\mmu) = \bb{g}'(\hat{\bb{x}})^{+}\bb{f}\left(t,\bb{g}(\hat{\bb{x}}),\mmu\right), \qquad \tilde{\bb{x}}(0,\mmu) = \tilde{\bb{x}}_0(\mmu) = \bb{g}\left(\hat{\bb{x}}(0,\mmu)\right),
\end{equation}
where $(\bb{g}')^+ = (\bb{g}'^\top\bb{g}')^{-1}\bb{g}'^\top$ denotes the Moore-Penrose pseudoinverse of $\bb{g}'$.
This is a well posed low-dimensional dynamical system that is feasible to solve when $\bb{g}$ and its derivatives are known.  Note that the ROM \eqref{eq:rom} takes place on the tangent bundle to the solution manifold $\mathcal{S}$, so that $\hat{\bb{x}}$ is a low-dimensional integral curve in this space.  Moreover, it is clear that choosing the mapping $\bb{g} = \bb{U}_n$ immediately recovers the standard POD-ROM from before since $\bb{U}_n$ is a linear mapping.


Solving the ROM \eqref{eq:rom} is often accomplished using Newton or quasi-Newton methods based on minimizing the residual (e.g.  \cite{lee2020,fu2018}), though recent work in \cite{eivazi2020,fresca2021} has demonstrated that a purely data-driven approach can also be effective for end-to-end reduced-order modeling of PDEs without appealing to conventional numerical solvers.  More precisely, instead of relying on a low-dimensional system of differential equations to compute the latent representation $(t,\mmu) \mapsto \hat{\bb{x}}(t,\mmu)$, it is reasonable to consider computing this quantity with another neural network which is fully connected and relatively simple.  In this case, the autoencoder $\bb{g}\circ\bb{h}$ can be trained alongside the mapping $\hat{\bb{x}}$ through the joint loss
\begin{equation}\label{eq:romloss}
    L(\bb{x},t,\mmu) = \| \bb{x} - \bb{g}\circ\bb{h}(\bb{x}) \|^2 + \| \hat{\bb{x}}(t,\mmu) - \bb{h}(\bb{x})\|^2.
\end{equation}
The first term in $L$ is simply the reconstruction error coming from the autoencoder, while the second term encourages the latent representation $\hat{\bb{x}}$ to agree with the encoded state $\bb{h}(\bb{x})$.  This approach has the advantage of requiring only function evaluations in its online stage, at the potential cost of model generalizability.  On the other hand, experiments in \cite{fresca2021} as well as Section~\ref{sec:numerics} of this work show that data-driven ROMs still produce competitive results in practice while in some cases maintaining a lower online cost.

\begin{remark}
\rev{
Alternatively to using the joint loss \eqref{eq:romloss}, its composite terms can be used separately to train the mapping $\hat{\bb{x}}$ after the mappings $\bb{g},\bb{h}$ in a two-stage process.  However, in practice we observe similar to \cite{fresca2021} that network ROM performance is not significantly affected by this choice.  This could be due to the observed accuracy bottleneck coming from $\hat{\bb{x}}$ (c.f. Section~\ref{sec:numerics}).
}
\end{remark}




It is important to keep in mind that the existence of the decoder mapping $\bb{g}$ is independent of the numerical technique used to generate it, and therefore the use of the convolutional propagation rule \eqref{eq:cnn} is a design decision rather than a necessity.  In fact, at present there are an enormous number of different neural network architectures available for accomplishing both predictive and generative tasks, each with their own unique properties.  Therefore, it is reasonable to question whether or not deep CAEs based on \eqref{eq:cnn} provide the best choice of network architecture for ROM applications.  The goal of the sequel is to introduce an alternative to the conventional CAE, along with some experiments which compare the performance of this CAE-ROM to other autoencoder-based ROMs involving fully connected and graph convolutional neural networks.
\section{A Graph Convolutional Autoencoder for ROM}
The convolutional propagation rule \eqref{eq:cnn} comes with inherent advantages and disadvantages that become readily apparent in a ROM setting.  One very positive quality of this rule is that if the convolutional filter is of size $K\times K$, then its associated discrete convolution is precisely $K$-localized in space.  This locality ensures that only the features of nearby nodes are convolved together, increasing the interpretability and expressiveness of the neural network.  On the other hand, it is clear that \eqref{eq:cnn} does not translate immediately to irregular domains, where each node has a variable number of neighbors in its $K$-hop neighborhood.  Because of this, bothersome tricks are needed to apply the CAE based on \eqref{eq:cnn} to PDE data defined on irregular or unstructured domains.  In particular, one popular approach (c.f. \cite{fresca2021}) is to simply pad and reshape the solution.  More precisely, each feature of $\bb{x}$ can be flattened into a vector, zero-padded until it has length $4^m$ for some $m$, and then reshaped to a square of size $2^m \times 2^m$.  This square is then fed into the CAE, generating an output of the same shape which can be flattened and truncated to produce the required feature of $\tilde{\bb{x}}$.  While this procedure works surprisingly well in many cases, it has the potential to create a very high overhead cost due to artificial padding of the solution.  Moreover, it fails to preserve the beneficial locality property of convolution, since nodes which were originally far apart may end up close together after reshaping.  Remedying these issues requires a new autoencoder architecture based on graph convolutional neural networks, which will now be discussed.



\subsection{Graph Convolutional Neural Networks}

While the discrete convolution in \eqref{eq:cnn} is in some sense a direct translation of the continuous version, it is only meaningful for regular, structured data.  However, PDE simulations based on high-fidelity numerical techniques such as finite element methods frequently involve domains which are irregular and unstructured.  In this case, developing an autoencoder-based ROM requires a notion of convolution which is appropriate for data with arbitrary connectivity.  On the other hand, it is not immediately obvious how to define a useful notion of convolution when the domain in question is unstructured.  Because discrete procedures typically obey a ``no free lunch'' scenario with respect to their continuous counterparts, it is an active area of research to determine which properties of the continuous convolution are most beneficial when preserved in the discrete setting.  Since unstructured domains can be effectively modeled as mathematical graphs, this has motivated a large amount of activity in the area of graph convolutional neural networks (GCNNs).  The next goal is to recall the basics of graph convolution and propose an autoencoder architecture which makes use of this technology.




To that end, consider an undirected graph $\mathcal{G} = \{\mathcal{V}, \mathcal{E}\}$ with (weighted or unweighted) adjacency matrix $\bb{A} \in \mathbb{R}^{|\mathcal{V}| \times |\mathcal{V}|}$.  Recall that $\bb{A} = (a_{ij})$ is symmetric with entries defined by 
\[ 
a_{ij} =
    \begin{cases} 
      w_{ij} & \exists\, e_{ij}\in\mathcal{E}\,\,\mathrm{connecting\,}v_i\mathrm{\,and\,}v_j \\
      0 & \mathrm{otherwise}, \\
    \end{cases}
\]
where $w_{ij}>0$ are the weight values associated to each edge.  Given this information, there are a wide variety of approaches to defining a meaningful convolution between two signals $f,g:\mathcal{V} \to \mathbb{R}^n$ defined at the nodes of $\mathcal{G}$ (see e.g. \cite{wu2020}).  One highly successful approach is to make use of the fact that convolution is just ordinary multiplication in Fourier space.  In particular, recall the combinatorial Laplacian $\bb{L}:\mathbb{R}^{|\mathcal{V}|}\to \mathbb{R}^{|\mathcal{V}|}$ which can be computed as $\bb{L} = \bb{D}-\bb{A}$ where $\bb{D} = (d_{ij})$ is the diagonal degree matrix with entries $d_{ii} = \sum_j a_{ij}$.  It is straightforward to check that $\bb{L}$ is symmetric and positive semi-definite (see e.g. \cite{bollobas2013}), therefore $\bb{L} = \bb{U}\bb{\Lambda}\bb{U}^\top$ has a complete set of real eigenvalues $\lambda_i$ with associated orthonormal eigenvectors $\bb{u}_i$ known as the Fourier modes of $\mathcal{G}$.  Since any signal $f:\mathcal{V} \to \mathbb{R}^n$ can be considered an $n$-length array of $|\mathcal{V}|$-vectors denoted $\bb{f} \in \mathbb{R}^{n\times|\mathcal{V}|}$, this yields the discrete Fourier transform $\hat{\bb{f}}_j = \bb{U}^\top \bb{f}_j$ where $\bb{f}_j \in \mathbb{R}^{|\mathcal{V}|}$ is the $j^{th}$ vectorized component function of $f$, and each entry $\hat{f}_j(\lambda_i) = \IP{}{\bb{u}_i}{\bb{f}_j}$.  The advantage of this notion is a straightforward extension of convolution to graphical signals $f,g:\mathcal{V} \to \mathbb{R}^n$ through multiplication in frequency space.  In particular, it follows that for $1\leq i,j\leq n$
\[ \bb{f}_i \ast \bb{g}_j = \bb{U}\left( \bb{U}^\top\bb{f}_i \odot \bb{U}^\top\bb{g}_j \right), \]
where $\odot$ denotes the element-wise Hadamard product and $\bb{f}_i = \bb{U}\hat{\bb{f}}_i$ is the inverse Fourier transform on $\mathcal{G}$.  Note the usual convention that matrix products occur before the Hadamard product.  


This expression provides a mathematically precise (though indirect) definition of spatial graph convolution through the graph Fourier transform, but carries the significant drawback of requiring matrix multiplication with the Fourier basis $\bb{U}$ which is generically non-sparse.  To remedy this, Defferrard et al. \cite{defferrard2016} proposed considering matrix-valued filters which can be recursively computed from the Laplacian $\bb{L}$.  In particular, consider a convolution kernel $g_\theta(\bb{L}) = \bb{U}g_\theta(\bb{\Lambda})\bb{U}^T$ where
$g_\theta(\bb{\Lambda}) = \sum_{k=0}^K \theta_k\bb{\Lambda}^k$ is an order $K$ polynomial in the eigenvalues of the \textit{normalized} Laplacian $\bb{L} = \bb{I} - \bb{D}^{-1/2}\bb{A}\bb{D}^{-1/2}$ and $\theta$ is a vector of learnable coefficient parameters.  Such kernels are easily computed if the eigenvalues $\bb{\Lambda}$ are known, but do not obviously eliminate the need for multiplication by the Fourier basis $\bb{U}$.  On the other hand, if $g_\theta(\bb{L})$ can be computed recursively from $\bb{L}$, then for any signal $\bb{x}\in\mathbb{R}^{|V|}$ the convolution $g_\theta \ast \bb{x} = g_\theta(\bb{L})\bb{x}$ can be computed in $\mathcal{O}(K|\mathcal{E}|) \ll \mathcal{O}(|\mathcal{V}|^2)$ operations, making the spectral approach more feasible for practical neural network applications.  As an added benefit of the work in \cite{defferrard2016}, note that order-$K$ polynomial kernels are also precisely $K$-localized in space.  In particular, the $j^{th}$ component of the kernel $g_\theta$ localized at vertex $i$ is given by
\[ \left(g_\theta(\bb{L})\bm{\delta}_i\right)_j = \left(g_\theta(\bb{L})\right)_{ij} = \sum_{k=0}^K \theta_k (\bb{L}^k)_{ij}, \]
where $\bm{\delta}_i$ is the $i^{th}$ nodal basis vector and $(\bb{L}^k)_{ij} = 0$ whenever $v_j$ lies outside the $K$-hop neighborhood of $v_i$ (c.f. \cite[Lemma 5.2]{hammond2011}). This shows that the frequency-based convolution $g_\theta(\bb{L})\bb{x}$ maintains the desirable property of spatial locality when the filter is a polynomial in $\bb{L}$.  

Further simplification to this idea was introduced by Kipf and Welling \cite{kipf2016} who chose $K=1, \theta_0 = 2\theta$, and $\theta_1 = -\theta$ with $\theta$ a learnable parameter to obtain the first-order expression $g_\theta(\bb{L})\bb{x} = \theta\left(\bb{I} + \bb{D}^{-1/2}\bb{A}\bb{D}^{-1/2}\right)\bb{x}$.  By renormalizing the initial graph to include self-loops (i.e. adding identity to the adjacency and degree matrices), they proposed the graph convolutional network (GCN) which obeys the propagation rule
\begin{equation}\label{eq:gcn1}
    \bb{x}_{\ell+1} = \sig\left(\tilde{\bb{P}}\bb{x}_{\ell}\bb{W}_\ell\right),
\end{equation}
where $\bb{x}_\ell \in \mathbb{R}^{|\mathcal{V}|\times n_\ell}$ is a signal on the graph with $n_\ell$ channels, $\bb{W}_\ell \in \mathbb{R}^{n_\ell \times n_{\ell+1}}$ is a (potentially nonsquare) weight matrix containing the learnable parameters and 
\[\tilde{\bb{P}} = \tilde{\bb{D}}^{-1/2}\tilde{\bb{A}}\tilde{\bb{D}}^{-1/2} = (\bb{D} + \bb{I})^{-1/2}(\bb{A} + \bb{I})(\bb{D} + \bb{I})^{-1/2}.\]
Networks of this form are simple and fast to compute, but were later shown \cite{wu2019} to be only as expressive as a fixed polynomial filter $g(\bm{\Lambda}) = (\bb{I} - \bm{\Lambda})^K$ on the spectral domain of the self-looped graph with order equal to the depth $K$ of the network.  While the propagation rule \eqref{eq:gcn1} displayed good performance on small-scale graphs,
it is now known that simple GCNs are notorious for overfitting and oversmoothing the data, as their atomic operation is related to heat diffusion on the domain.  This severely limits their expressive power, as GCNNs based on \eqref{eq:gcn1} are depth-limited in practice and cannot effectively reconstruct complex signals when used in a ROM setting.  

On the other hand,  many other graph convolutional operations are available at present, some of which were developed specifically to remedy this aspect of the original GCN.  One such operation was introduced in \cite{chen2020} and directly extends approach of \cite{kipf2016}.  In particular, the authors of \cite{chen2020} define the ``GCNII'' or GCN2 network, whose propagation rule is
\begin{equation}\label{eq:gcn2}
    \bb{x}_{\ell+1} = \sig\left[ \left( (1-\alpha_\ell)\tilde{\bb{P}}\bb{x}_\ell + \alpha_\ell\bb{x}_0 \right) \left( (1-\beta_\ell)\bb{I} + \beta_\ell\bb{W}_\ell \right) \right],
\end{equation}
where $\alpha_\ell, \beta_\ell$ are hyperparameters and $\sig$ is the component-wise ReLU function $\sigma(x) = \max\{x,0\}$.  The terms in this operation are motivated by the ideas of residual connection and identity mapping popularized in the well known ResNet architecture (c.f. \cite{he2016}), and the authors prove that $K$-layer networks of this form can express an arbitrary $K^{th}$-order polynomial in the self-looped graph Laplacian.  While the hyperparameters $\alpha_\ell$ are typically constant in practice, it is noted in \cite{chen2020} that improved performance is attained when $\beta_\ell$ decreases with increasing network depth.  Therefore, the experiments in Section~\ref{sec:numerics} use the recommended values of $\alpha_\ell = 0.2$ and $\beta_\ell = \log(1 + \frac{\theta}{\ell})$ where $\theta = 1.5$ is a constant hyperparameter.

With this, it is now possible to use the propagation rule \eqref{eq:gcn2} as the basis for a convolutional autoencoder.  Although spectral graph convolution does not produce any natural downsampling analogous to the CNN case, it is still reasonable to expect that a succession of convolutional layers will learn to extract important features of the data it accepts as input.  Therefore, it should be possible to recover the decoder mapping $\bb{g}$ necessary for ROM from this approach as well.  Moreover, in contrast to the discrete convolution \eqref{eq:cnn} the GCN2 operation is invariant to label permutation and remains $K$-localized on unstructured data.  Therefore, it is reasonable to suspect that a CAE-ROM based on \eqref{eq:gcn2} will exhibit superior performance when the PDE to be simulated takes place on an unstructured domain.  The remainder of the work will investigate whether or not this is the case, beginning with a description of the GCNN autoencoder that will be used.

\begin{remark}
Note that there are a wide variety of graph convolutional operations which can be used as the basis for a convolutional autoencoder on unstructured data.  During the course of this work, many such operations were invesigated including the ones proposed in \cite{bresson2017,gilmer2017,qi2017,ma2019, wang2019}.  Through experimentation, it was determined that the graph convolution proposed in \cite{chen2020} is the most effective for the present purpose.
\end{remark}

\subsection{Graph Convolutional Autoencoder Architecture}

Recall that the standard CAE is structured so that each layer downsamples the input in space at the same rate that it extrudes the input in the channel dimension.  In particular, it is common to use $5\times 5$ convolution kernels with a stride of two and ``same'' zero padding (c.f. \cite[Section 4.4]{dumoulin2016}), so that the spatial dimension of $\bb{x}$ decreases by a factor of four at each layer.  While this occurs, the channel dimension is simultaneously increased by a factor of four, so that the total dimensionality of $\bb{x}$ remains constant until the last layer(s) of the encoder.  The final part of the encoder is often fully connected, taking the simplified output of the convolutional network and mapping it directly to the encoded representation $\bb{h}(\bb{x})$ in some small latent space $\mathbb{R}^n$.  To recover $\tilde{\bb{x}} \approx \bb{x}$ from its encoding, this entire block of layers is simply ``transposed'' to form the decoder mapping $\bb{g}$, so that the autoencoder comprises the mapping $\bb{g}\circ\bb{h}$ as desired.  The learnable parameters of the mappings $\bb{g},\bb{h}$ are occasionally shared (resulting in a ``tied autoencoder''), though most frequently they are not.

\begin{figure}
    \centering
    \includegraphics[width=0.8\linewidth]{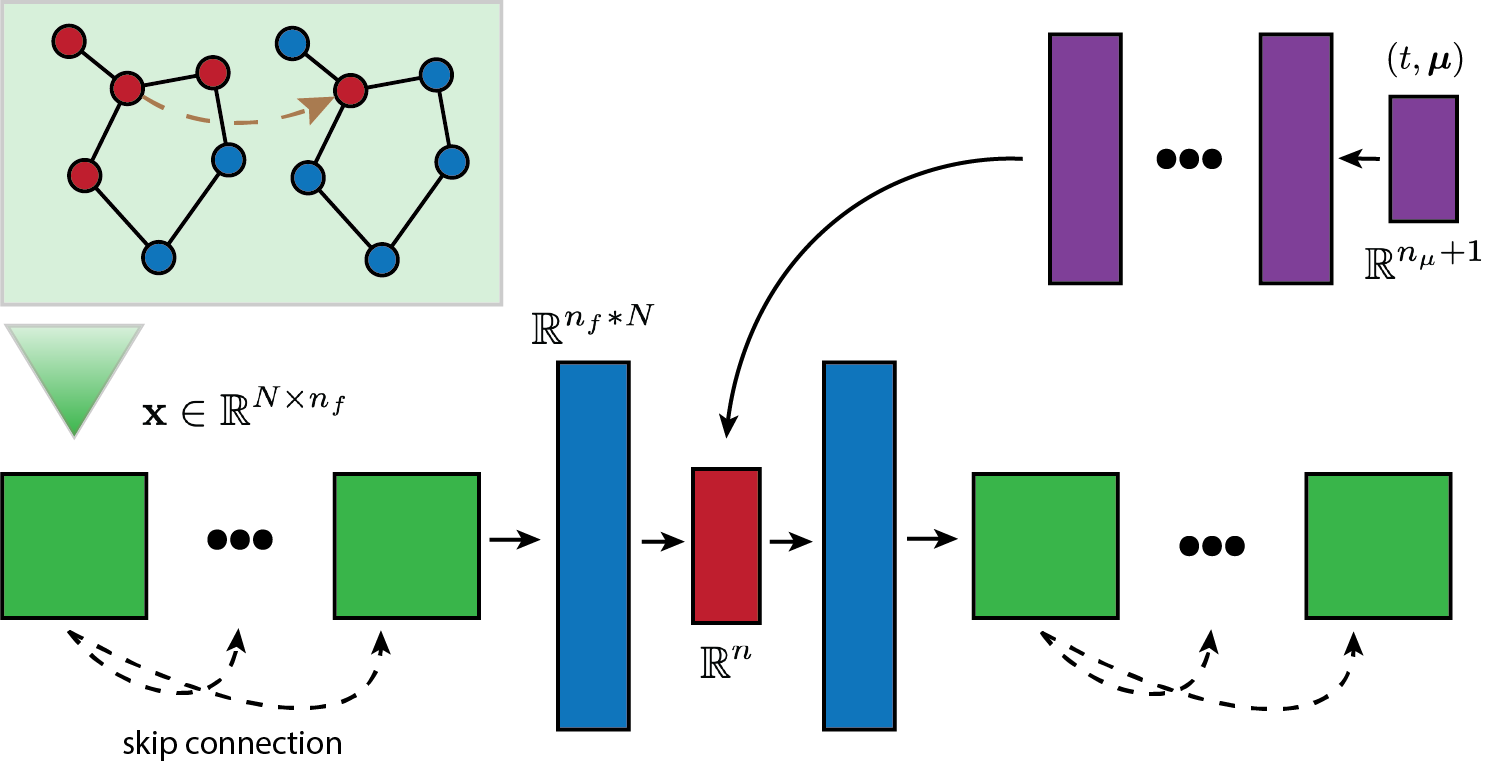}
    \caption{Illustration of the ROM architecture based on GCN2 graph convolution.}
    \label{fig:gcn2rom}
\end{figure}

To study whether autoencoder-based ROMs benefit from the core ideas behind GCNNs, we propose the ROM architecture displayed in Figure~\ref{fig:gcn2rom}.  Similar to \cite{fresca2021}, this ROM is entirely data-driven and is comprised of two neural networks: a specially designed CAE based on the operation \eqref{eq:gcn2} whose decoder is the required mapping $\bb{g}$, and a small-scale FCNN which simulates the low-dimensional mapping $(t,\mmu) \mapsto \hat{\bb{x}}(t,\mmu)$.  In this way, the component networks are trained directly from snapshot data during the offline stage, so that the online stage requires only function evaluations.  Recall that the input $\bb{x}$ is not downsampled by the convolutional procedure \eqref{eq:gcn2}, so the feature dimension is similarly held constant.  Moreover, use of the GCN2 operation ensures that each convolutional layer is skip-connected to the first, i.e. each convolutional layer retains a fraction of the pre-convolutional input.  Therefore, the proposed autoencoder is defended against the troublesome oversmoothing known to occur in deep GCNs.  To evaluate the performance of our GCNN-ROM, it is compared directly to two analogous autoencoder-based data-driven ROMs which use the same prediction network $\hat{\bb{x}}$ and loss function \eqref{eq:romloss} but different autoencoder architectures.  The first is the standard deep CAE based on \eqref{eq:cnn} and designed according to the common conventions of CAEs on regular data, and the second is a simple feed-forward autoencoder which uses batch-normalized fully connected layers throughout.  The results of a comparison between these ROMs are given in the next Section.


\section{Numerical Examples}\label{sec:numerics}
This section details a performance comparison between ROMs using autoencoder architectures based on fully connected (FCNN), convolutional (CNN), and graph convolutional (GCNN) networks.  Through examples of PDEs on structured and unstructured domains, ROM performance is evaluated on two distinct tasks.  The first task measures how well each architecture can approximate PDE solutions at new parameter values (referred to as the \emph{prediction problem}), and the second measures how well they can compress and decompress PDE data (referred to as the \emph{compression problem}).



\subsection{Data Preprocessing and Evaluation}
Begin with a snapshot array $\bb{S} \in \mathbb{R}^{N_s \times N \times n_f}$ each row of which is a solution $\bb{x}(t,\mmu)$ at a particular parameter configuration $(t,\mmu)$ found in the (consistently ordered) parameter matrix $\bb{M}\in\mathbb{R}^{N_s \times (n_\mu+1)}$.  We first choose integers $N_t, N_v$ such that $N_s = N_t + N_v$ and separate $\bb{M} = (\bb{M}_t, \bb{M}_v)$ and $\bb{S} = (\bb{S}_t, \bb{S}_v)$ row-wise into training and validation sets, taking care that no value of $\mmu$ appears in both $\bb{M}_t, \bb{M}_v$.  After separation, the training data $\bb{M}_t, \bb{S}_t$ is normalized ``column-wise'' to the interval $[0,1]$.  Momentarily dropping the subscript $t$, this means computing the transformation
\begin{equation*}
    \bb{M}^s \mapsto \frac{\bb{M}^s - \bb{M}_{\mathrm{min}}}{\bb{M}_{\mathrm{max}} - \bb{M}_{\mathrm{min}}} \qquad     \bb{S}^s \mapsto \frac{\bb{S}^s - \bb{S}_{\mathrm{min}}}{\bb{S}_{\mathrm{max}} - \bb{S}_{\mathrm{min}}},
\end{equation*}
where $\bb{M}_{\mathrm{max}}, \bb{M}_{\mathrm{min}}$ (resp. $\bb{S}_{\mathrm{max}}, \bb{S}_{\mathrm{min}}$) are arrays containing the element-wise maxima and minima of $\bb{M}$ (resp. $\bb{S})$ over the sampling dimension $1\leq s \leq N_t$ (e.g. $\left(M_{\mathrm{max}}\right)_{ij} = \max_s M^s_{ij}$ for every relevant $i,j$), and the division is performed element-wise.  Once the training data has been normalized, the validation data $\bb{M}_v, \bb{S}_v$ is normalized in a similar fashion based on the arrays $\bb{M}_{max/min},\bb{S}_{max/min}$ computed from the training set.


Once the relevant neural networks have been trained, performance is measured through the empirical relative $\ell_1, \ell_2$ reconstruction of the FOM solution $\bb{x}$ over the validation set.  In particular, denoting the reconstructed solution by $\tilde{\bb{x}}$, the empirical relative reconstruction error is computed as
\begin{equation*}
    R\ell_i(\bb{x},\tilde{\bb{x}}) = \frac{\sum_s\nn{\bb{x}^s - \tilde{\bb{x}}^s}_i}{\sum_s\nn{\bb{x}^s}_i},
\end{equation*}
where $1\leq s \leq N_v$.
Note that the reconstructed solution $\tilde{\bb{x}} = \bb{g}\circ\bb{h}(\bb{x})$ for the compression problem while $\tilde{\bb{x}} = \bb{g}\circ\hat{\bb{x}}(t,\mmu)$ for the prediction problem.  In addition to the reconstruction error, it is helpful to report the memory required to store the trained models as well as the average wall-clock time per training epoch.  While the memory cost is a property of the network, note that the wall-clock time is hardware and implementation dependent with significant fluctuation throughout the course of training.  On the other hand, it  provides a useful comparison (along with the plots of the loss) to estimate the difference in computational cost across autoencoder architectures.  Pseudocode for the training procedure is given in Algorithm~\ref{alg:training}.  Note that an early stopping criterion is employed, so that training is stopped when the validation loss does not decrease over a predefined number of consecutive epochs.  All experiments are performed using Python 3.8.10 running PyTorch 1.8.1 \cite{torch} on an early 2015 model MacBook Pro with 2.7 GHz Dual-Core Intel Core i5 processor and 8GB of RAM.  The implementation of GCN2 convolution is modified from source code found in the PyTorch Geometric package \cite{fey2019}. 


\begin{algorithm}
\SetAlgoLined
\KwResult{Optimized functions $\bb{g},\bb{h}, \hat{\bb{x}}$. }
Initialize: normalized arrays $\bb{M}_t\in \mathbb{R}^{N_t \times (n_\mu + 1)}$, $\bb{S}_t\in \mathbb{R}^{N_t \times N \times n_f}$, $\bb{M}_v\in \mathbb{R}^{N_v \times (n_\mu + 1)}$, $\bb{S}_v\in \mathbb{R}^{N_v \times N \times n_f}$, early stopping parameter $c\in\mathbb{N}$, $N_B, n_b$\;
$\mathrm{count}=0$ \tcp{stopping criterion}
$\mathrm{loss}=100$ \tcp{comparison criterion}
\While{$\mathrm{count} < c$}{
Randomly (but consistently) shuffle $\bb{M}_t, \bb{S}_t$ into $N_B$ batches $\bb{M}_{t,b}, \bb{S}_{t,b}$ of size $n_b$\;
    \For{$1\leq \mathrm{batch} \leq N_B$}{
        Compute the loss $L(\bb{x},t,\mmu)$ over the $n_b$ samples $\bb{x}\in\bb{S}_{t,b}, (t,\mmu)\in\bb{M}_{t,b}$\;
        Backpropagate corresponding gradient information\; 
        Update network parameters based on ADAM rules\;
    }
    Compute and return $L_v = L(\bb{x},t,\mmu)$ over the $N_v$ samples $\bb{x}\in\bb{S}_{v}, (t,\mmu)\in\bb{M}_{v}$\;
    \uIf{ $L_v < \mathrm{loss}$}{
        $\mathrm{loss} = L_v$\;
        $\mathrm{count} = 0$\;
    }
    \Else{count++}
}
\caption{Network Training}
\label{alg:training}
\end{algorithm}


\begin{table}[]
\bgroup
\def\arraystretch{1.25}%
\begin{tabular}{|c|c|c|c|c|c|c|}
\hline
\textbf{layer} & \textbf{input size} & \textbf{kernel size} & \textbf{$\alpha$} & \textbf{$\theta$} & \textbf{output size} & \textbf{activation} \\ \hline
1-C & $(N, n_f)$ & $n_f \times n_f$ & 0.2 & 1.5 & $(N, n_f)$ & ReLU \\ \hline
\multicolumn{7}{|c|}{......} \\ \hline
$N_l$-C & $(N, n_f)$ & $n_f \times n_f$ & 0.2 & 1.5 & $(N, n_f)$ & ReLU \\ \hline
\multicolumn{7}{|c|}{Samples of size $(N, n_f)$ are flattened to size $(n_f*N)$.} \\ \hline
1-FC & $n_f*N$ & \multicolumn{3}{c|}{} & $n$ & Identity \\ \hline
\multicolumn{7}{|c|}{End of encoding layers.  Beginning of decoding layers.} \\ \hline
2-FC & $n$ & \multicolumn{3}{c|}{} & $n_f*N$ & Identity \\ \hline
\multicolumn{7}{|c|}{Samples of size $(n_f*N)$ are reshaped to size $(N, n_f)$.} \\ \hline
1-TC & $(N, n_f)$ & $n_f \times n_f$ & 0.2 & 1.5 & $(N, n_f)$ & ReLU \\ \hline
\multicolumn{7}{|c|}{......} \\ \hline
$N_l$-TC & $(N, n_f)$ & $n_f \times n_f$ & 0.2 & 1.5 & $(N, n_f)$ & ReLU \\ \hline
\multicolumn{7}{|c|}{End of decoding layers.  Prediction layers listed below.} \\ \hline
3-FC & $n_\mu+1$ & \multicolumn{3}{c|}{} & 50 & ELU \\
4-FC & 50 & \multicolumn{3}{c|}{} & 50 & ELU \\ \hline
\multicolumn{7}{|c|}{......} \\ \hline
10-FC & 50 & \multicolumn{3}{c|}{} & $n$ & Identity \\ \hline
\end{tabular}
\egroup
\bigskip
\caption{Architecture of GCNN-based autoencoder.  Note that $\beta_\ell = \log\left(1 + \frac{\theta}{\ell}\right)$ for the C layers and $\beta_\ell = \log\left(1 + \frac{\theta}{N_l + \ell}\right)$ for the TC layers.}
\label{tab:GCNN}
\end{table}

\subsection{Parameterized 1-D Inviscid Burger's Equation}
The first experiment considers the parameterized one-dimensional inviscid Burgers' equation, which is a common model for fluids whose motion can develop discontinuities.  In particular, let $\mmu = \left(\mu_1\,\,\mu_2\right)^\top$ be a vector of parameters and consider the initial value problem,
\begin{align}\label{eq:burgers}
\begin{split}
    w_t + \frac{1}{2}\left(w^2\right)_x &= 0.02 e^{\mu_2 x}, \\
    w(a, t, \mmu) &= \mu_1, \\
    w(x,0,\mmu) &= 1,
\end{split}
\end{align}
where subscripts denote partial differentiation and $w=w(x,t, \mmu)$ represents the conserved quantity of the fluid for $x\in[a,b]$.  As a first test, it is interesting to examine how the CNN, GCNN, and FCNN autoencoder-based ROMs perform on this relatively simple hyperbolic equation \rev{when compared to linear techniques such as POD.}  First, note that this problem is easily converted to the form \eqref{eq:fom} by discretizing the interval $[a,b]$ with finite differences, in which case the solution at each nodal point is $w^i(t,\mmu) \coloneqq w(x_i,t,\mmu)$, and $\bb{w}(t,\mmu) \in \mathbb{R}^N$ is a vector for each $(t,\mmu)$ of length equal to the number $N$ of nodes.  With this, the equations \eqref{eq:burgers} can be discretized and solved for various parameter values using a simple upwind forward Euler scheme.  In particular, given stepsizes $\Delta x, \Delta t$, let $w^i_k$ denote the solution $\bb{w}$ at time step $k$ and spatial node $i$.  Then, simple forward differencing gives
\begin{align*}
    \frac{w_{k+1}^i - w_k^i}{\Delta t} + \frac{1}{2}\frac{\left(w^2\right)_k^i - \left(w^2\right)_k^{i-1}}{\Delta x} &= 0.02e^{\mu_2 x},
\end{align*}
which expresses the value $w^i$ at $t = (k+1)\Delta t$ in terms of its value at $t = k\Delta t$.  For the experiment shown, the discretization parameters chosen are $\Delta x = 100/256$ for $x \in [0,100]$ and $\Delta t = 1/10$ for $t\in[0,10]$.  The parameters $\mmu$ used for training are $\{(2 + 0.5i, 0.015 + 0.005j)\}$ where $0\leq i \leq 2$ and $0\leq j \leq 3$, while the validation parameters are the lattice midpoints $\{(2.25 + 0.5i, 0.0175 + 0.005j)\}$ where $0 \leq i \leq 1$ and $0\leq j\leq 2$.  A precise description of the autoencoder architectures employed can be found in Tables~\ref{tab:GCNN}, \ref{tab:burgFCNN}, and \ref{tab:burgCNN}, \rev{and a precise description of the POD-ROM employed can be found in Appendix~\ref{app:POD}}.  The initial learning rates for the ADAM optimizer are chosen as 0.0025 for in the GCNN case, 0.001 in the CNN case, and 0.0005 in the FCNN case.  The early stopping criterion for this example is 200 epochs and the mini-batch size is 20.

As seen in Table~\ref{tab:burgers}, the results for both the prediction and compression problems are significantly different across ROM architectures and latent space sizes.  Indeed, the CNN autoencoder is consistently the most accurate on the prediction problem when $n$ is small while also being the most time consuming and memory consumptive, and its performance does not improve with increased latent dimension.  \rev{This is consistent with prior work demonstrating that nonlinear ROMs can outperform linear techniques such as POD when $n$ is close to the theoretical minimum.  Conversely, the quality of the POD-ROM approximation continues to improve as more modes are added and eventually surpasses the performance of all network-based architectures.  This is not surprising, as the POD-ROM is not strictly data-driven like the neural network ROMs and so benefits from additional structure.}  Note that the performance of the GCNN-ROM increases quite a bit when the latent dimension is increased from 3 to 10, but remains significantly less accurate than the ROM based on CNN or FCNN. Interestingly, on the prediction problem the performance of all networks appears to be bottlenecked by the fully connected network $\hat{\bb{x}}$ which simulates the low-dimensional dynamics, as their accuracy does not increase when the latent dimension increases from $n=10$ to $n=32$.  This suggests that a ROM based on Newton or quasi-Newton methods may produce better results in this case.

On the compression problem, the results show that all ROM architectures benefit from an increased latent space dimension when the goal is to encode and decode, with the GCNN autoencoder benefiting the most and the CNN autoencoder benefiting the least.  Again, the GCNN architecture struggles when the latent space dimension is set to the theoretical minimum $n=3$, however its performance exceeds CNN and becomes competitive with that of FCNN once the latent dimension is increased to $n=32$, at half of the memory cost.  \rev{However, at this size the linear POD approximation is also quite good, nearly matching the neural network approximations in quality.}

\begin{remark}
Instead of increasing the latent dimension $n$, it is reasonable to consider increasing the depth of the autoencoder and prediction networks.  However, we find that increasing the depth of the GCNN autoencoder beyond $N_l = 4$ does not significantly improve its performance on either the prediction or compression problems.  On the other hand, the GCNN-ROM benefits modestly from an increase in depth of the prediction network from 4 to 8 layers, although the CNN and FCNN based ROMs do not.  Therefore, all prediction experiments use 8 fully connected layers in the GCNN case and 4 in the CNN and FCNN cases.
\end{remark}

\begin{figure}
    \centering
    \begin{minipage}{0.5\textwidth}
        \includegraphics[width=0.8\linewidth]{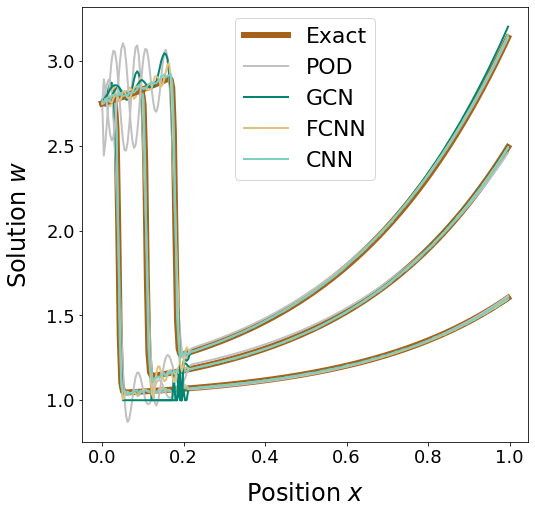}
    \end{minipage}%
    \begin{minipage}{0.5\textwidth}
        \includegraphics[width=0.8\linewidth]{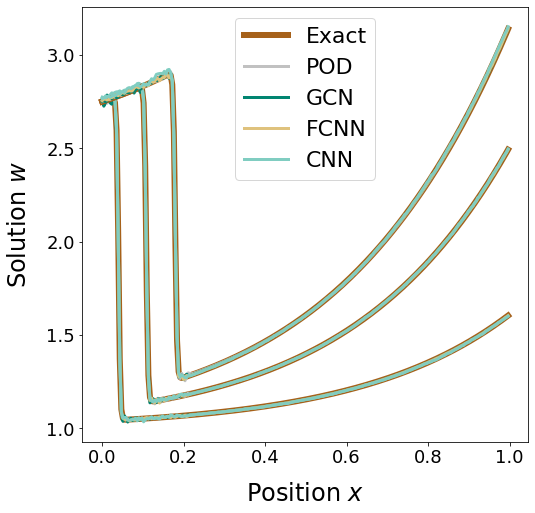}
    \end{minipage}
    \caption{\rev{Plots corresponding to the Burgers' example with $\mmu=\begin{pmatrix}0.275 & 0.0275\end{pmatrix}^\top$ and $t=0.4,1.1,1.8$.  Left: results for prediction problem with $n=10$.  Right: results for compression problem with $n=32$.}}
    \label{fig:burgersPlot}
\end{figure}

\begin{figure}
    \centering
    \begin{minipage}{0.5\textwidth}
        \includegraphics[width=\linewidth]{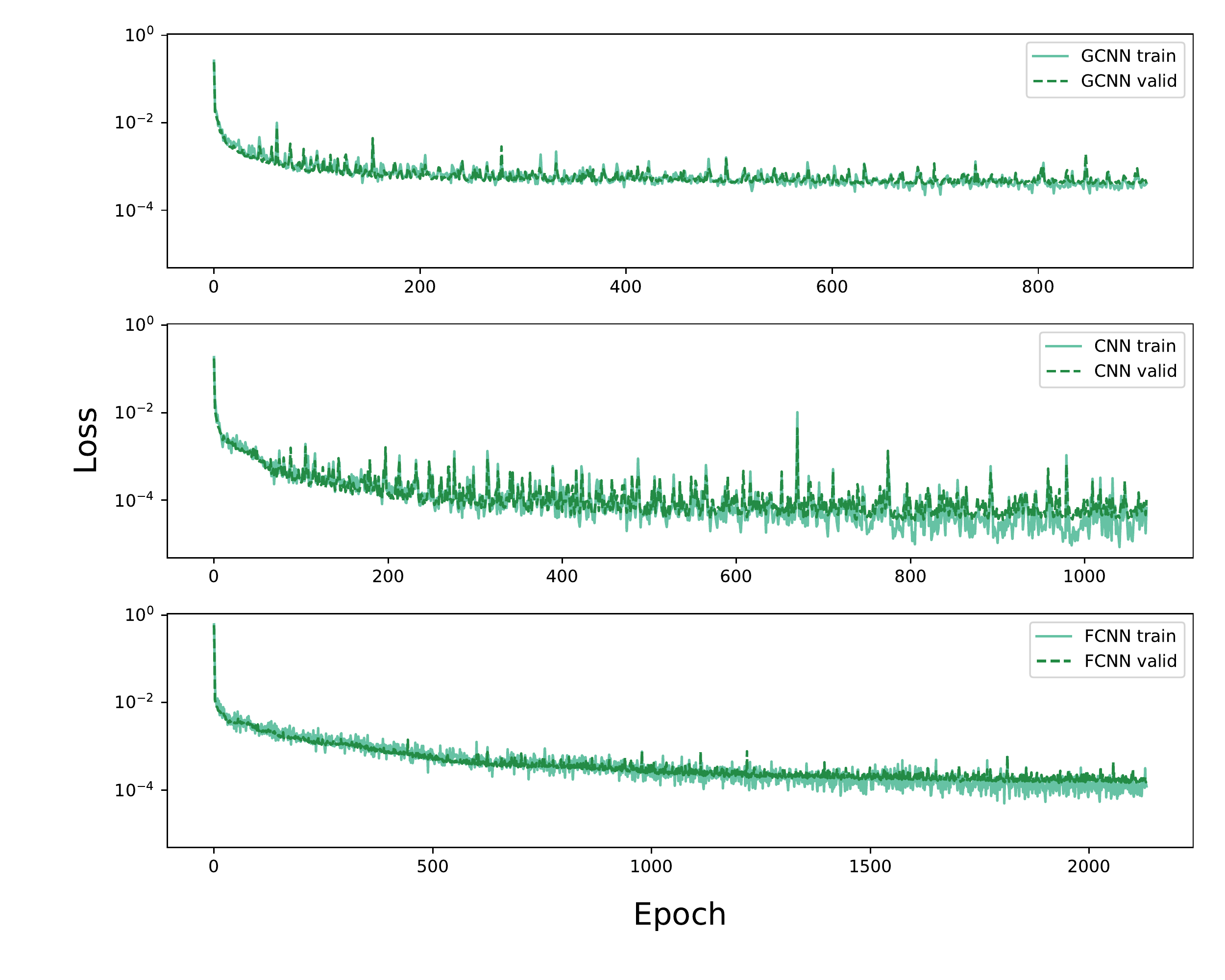}
    \end{minipage}%
    \begin{minipage}{0.5\textwidth}
        \includegraphics[width=\linewidth]{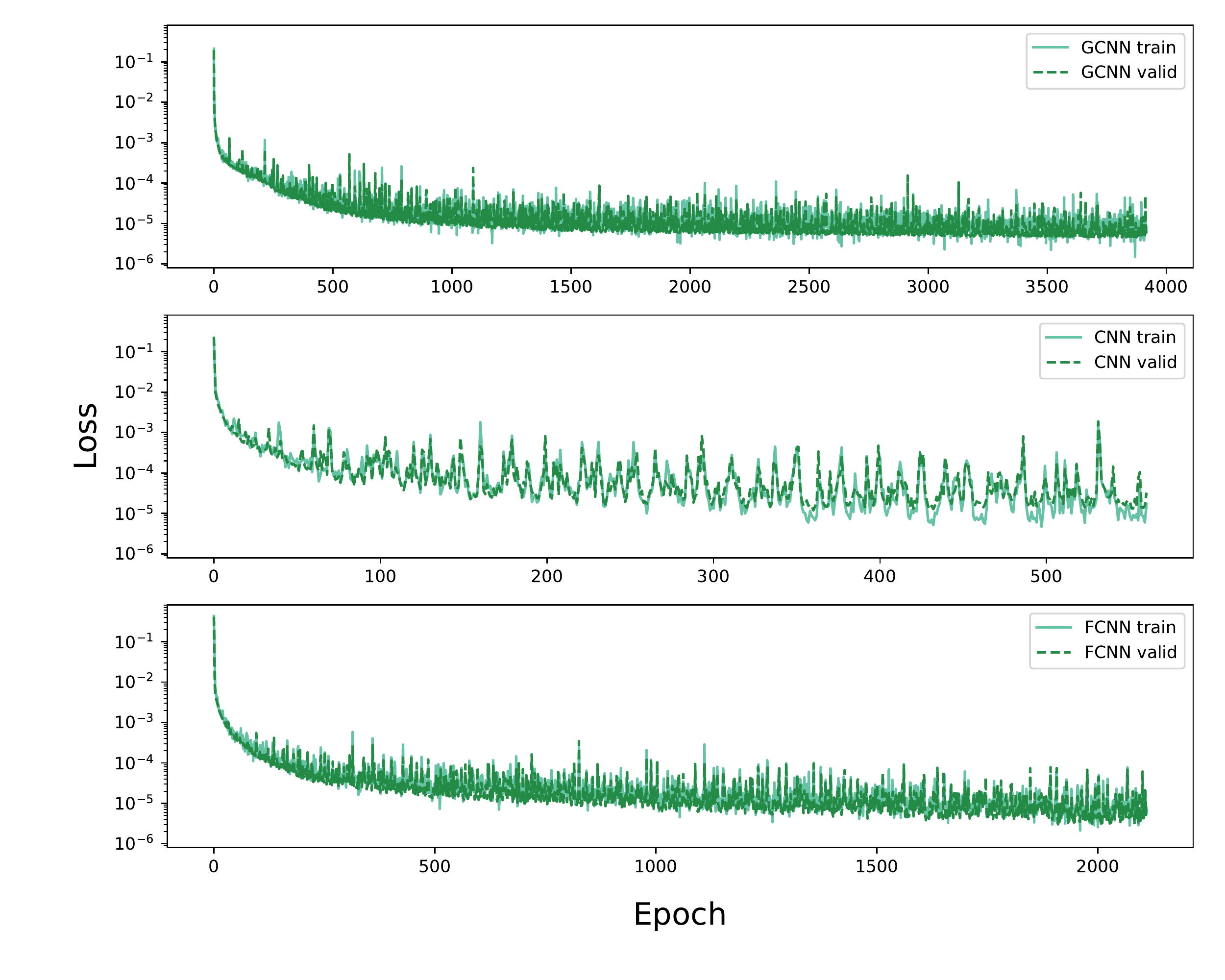}
    \end{minipage}
    \caption{Training and validation losses incurred during neural network training for the 1-D Burgers' example.  Left: prediction problem with $n=10$.  Right: compression problem with $n=32$.}
    \label{fig:burgersLoss}
\end{figure}

Figure~\ref{fig:burgersPlot} illustrates these results at various values of $t$ for the parameter $\mu=\begin{pmatrix}2.75 & 0.0275\end{pmatrix}^\top$.  Clearly, the architecture based on CNN is best equipped for solving the prediction problem in this case.  On the other hand, when $n=32$ it is evident that all architectures can compress and reconstruct solution data well, although the FCNN and GCNN yield the best results.  Figure~\ref{fig:burgersLoss} shows the corresponding plots of the losses incurred during network training.  Note that all architectures have comparable training and validation losses for the compression problem, while all but the GCNN overfit the data slightly on the prediction problem.

\begin{table}[]
\begin{tabular}{|c|c|c|c|}
\hline
\textbf{layer} & \textbf{input size} & \textbf{output size} & \textbf{activation} \\ \hline
\multicolumn{4}{|c|}{Samples of size (256,1) are flattened to size (256).} \\ \hline
1-FC & 256 & 64 & ELU \\ 
1-BN & 64 & 64 & Identity \\ 
2-FC & 64 & $n$ & ELU \\ \hline
\multicolumn{4}{|c|}{End of encoding layers.  Beginning of decoding layers.} \\ \hline
3-FC & $n$ & 64 & ELU \\ 
2-BN & 64 & 64 & Identity \\
4-FC & 64 & 256 & ELU \\ \hline
\multicolumn{4}{|c|}{Samples of size (256) are reshaped to size (256,1)} \\ \hline
\multicolumn{4}{|c|}{End of decoding layers.  Prediction layers listed below.} \\ \hline
5-FC & $n_\mu+1$ & 50 & ELU \\ 
6-FC & 50 & 50 & ELU \\ 
7-FC & 50 & 50 & ELU \\ 
8-FC & 50 & $n$ & ELU \\ \hline
\end{tabular}
\bigskip
\caption{FCNN architecture for the 1-D Burgers' example.}
\label{tab:burgFCNN}
\end{table}

\begin{table}[]
\begin{tabular}{|c|c|c|c|c|c|c|}
\hline
\textbf{layer} & \textbf{input size} & \textbf{kernel size} & \textbf{stride} & \textbf{padding} & \textbf{output size} & \textbf{activation} \\ \hline
\multicolumn{7}{|c|}{Samples of size (256,1) are reshaped to size (1, 16, 16).} \\ \hline
1-C & (1, 16, 16) & 5x5 & 1 & SAME & (8, 16, 16) & ELU \\
2-C & (8, 16, 16) & 5x5 & 2 & SAME & (16, 8, 8) & ELU \\
3-C & (16, 8, 8) & 5x5 & 2 & SAME & (32, 4, 4) & ELU \\
4-C & (32, 4, 4) & 5x5 & 2 & SAME & (64, 2, 2) & ELU \\ \hline
\multicolumn{7}{|c|}{Samples of size (64, 2, 2) are flattened to size (256).} \\ \hline
1-FC & 256 & \multicolumn{3}{c|}{} & $n$ & ELU \\ \hline
\multicolumn{7}{|c|}{End of encoding layers.  Beginning of decoding layers.} \\ \hline
2-FC & $n$ & \multicolumn{3}{c|}{} & 256 & ELU \\ \hline
\multicolumn{7}{|c|}{Samples of size (256) are reshaped to size (64, 2, 2).} \\ \hline
1-TC & (64, 2, 2) & 5x5 & 2 & SAME & (64, 4, 4) & ELU \\
2-TC & (64, 4, 4) & 5x5 & 2 & SAME & (32, 8, 8) & ELU \\
3-TC & (32, 8, 8) & 5x5 & 2 & SAME & (16, 16, 16) & ELU \\
4-TC & (16, 16, 16) & 5x5 & 1 & SAME & (1, 16, 16) & ELU \\ \hline
\multicolumn{7}{|c|}{Samples of size (1, 16, 16) are reshaped to size (256, 1).} \\ \hline
\end{tabular}
\bigskip
\caption{CNN autoencoder architecture for the 1-D Burgers' example, identical to \cite[Table 1,2]{fresca2021}.  Prediction layers (not pictured) are the same as in the FCNN case (c.f. Table~\ref{tab:burgFCNN}).}
\label{tab:burgCNN}
\end{table}

\begin{table}[h]
\begin{tabular}{|c|c|c|c|c|l|c|c|c|c|l|}
\hline
 & \multicolumn{5}{c|}{Encoder/Decoder $+$ Prediction} & \multicolumn{5}{c|}{Encoder/Decoder only} \\ \hline
Method & $n$ & $R\ell_1$\% & $R\ell_2$\% & Size (kB) & \begin{tabular}[c]{@{}l@{}}Time per\\ Epoch (s)\end{tabular} & $n$ & $R\ell_1$\% & $R\ell_2$\% & Size (kB) & \begin{tabular}[c]{@{}l@{}}Time per\\ Epoch (s)\end{tabular} \\ \hline
POD & \multirow{4}{*}{3} & 3.67 & 8.50 & 6.27 & N/A & \multirow{4}{*}{3} & 3.56 & 8.29 & 6.27 & N/A \\
GCNN &  & 4.41 & 8.49 & 79.0 & 0.55 &  & 2.54 & 5.31 & 13.1 & 0.46 \\
CNN &  & 0.304 & 0.605 & 965 & 2.2 &  & 0.290 & 0.563 & 953 & 2.2 \\
FCNN &  & 1.62 & 3.29 & 167 & 0.28 &  & 0.658 & 1.66 & 144 & 0.22 \\ \hline
POD & \multirow{4}{*}{10} & 1.75 & 3.63 & 20.6 & N/A & \multirow{4}{*}{10} & 1.18 & 2.86 & 20.6 & N/A \\
GCNN &  & 2.08 & 3.73 & 94.8 & 0.58 &  & 0.706 & 1.99 & 27.4 & 0.47 \\
CNN &  & 0.301 & 0.630 & 992 & 2.4 &  & 0.215 & 0.409 & 967 & 2.2 \\
FCNN &  & 0.449 & 1.15 & 172 & 0.27 &  & 0.171 & 0.361 & 148 & 0.23 \\ \hline
POD & \multirow{4}{*}{32} & 0.137 & 0.264 & 65.7 & N/A & \multirow{4}{*}{32} & 0.117 & 0.249 & 65.7 & N/A \\
GCNN &  & 2.59 & 4.17 & 145 & 0.59 &  & 0.087 & 0.278 & 72.6 & 0.46 \\
CNN &  & 0.350 & 0.675 & 1040 & 2.3 &  & 0.216 & 0.384 & 1010 & 2.2 \\
FCNN &  & 0.530 & 1.303 & 188 & 0.27 &  & 0.098 & 0.216 & 159 & 0.23 \\ \hline
\end{tabular}
\bigskip
\caption{\rev{Numerical results for the 1-D parameterized Burgers' example corresponding to the prediction problem (left) and the compression problem (right).}}
\label{tab:burgers}
\end{table}

\begin{figure}
    \centering
    \includegraphics[width=\linewidth]{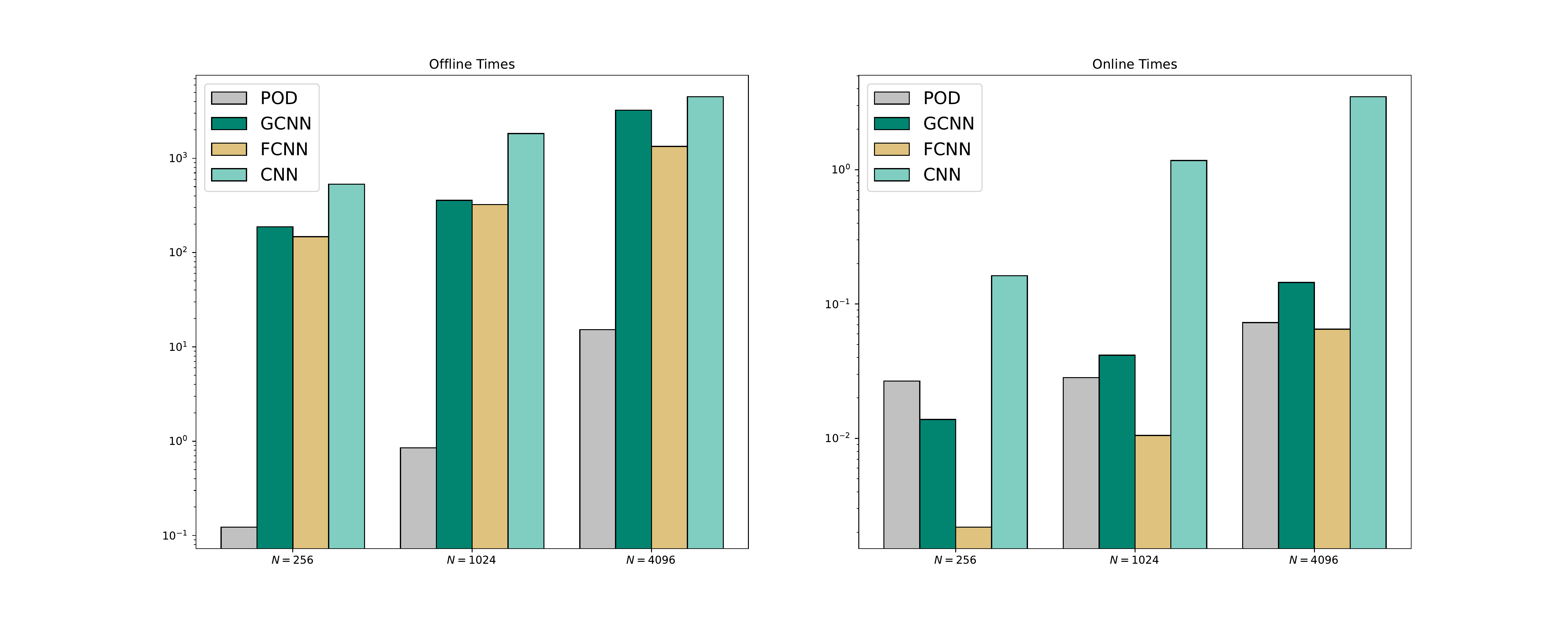}
    \caption{\rev{Plots of the computational costs in Table~\ref{tab:timings} corresponding to the 1-D parameterized Burgers' example.}}
    \label{fig:timings}
\end{figure}

\rev{It is also worthwhile to compare how the offline and online costs of each ROM architecture scale with increasing degrees of freedom.  Table~\ref{tab:timings} displays one such comparison using data from the present example.  Given the parameter $\mu=\begin{pmatrix}2.75 & 0.0275\end{pmatrix}^\top$, $N_s = 1200$ FOM snapshots are generated at the resolutions $N=256, 1024, 4096$, which are used to evaluate the computation time of each ROM in seconds.  The offline and online costs reported in Table~\ref{tab:timings} represent respectively the total time required to train each ROM and the time required to evaluate each trained ROM.  Note that $n=10$ is used, the offline cost of the POD-ROM includes the assembly time for the low-dimensional system, and the early stopping criterion for the neural networks is relaxed to 50 epochs.  Here and in Figure~\ref{fig:timings} it can be seen that none of the nonlinear ROMs can match the speed of POD offline, although both the GCNN and FCNN architectures are more efficient than POD online when the dimension of the problem is small enough.  Conversely, for this example POD scales better with increasing $n$ and eventually becomes more efficient in both categories.  Interestingly, the highest performing CNN architecture is seen to be the slowest both offline and online by a large margin, potentially because of its much larger memory size.}

\begin{table}[]
\begin{tabular}{|cc|cccccc|}
\hline
\multicolumn{2}{|c|}{\multirow{3}{*}{}} & \multicolumn{6}{c|}{N} \\ \cline{3-8} 
\multicolumn{2}{|c|}{} & \multicolumn{2}{c|}{256} & \multicolumn{2}{c|}{1024} & \multicolumn{2}{c|}{4096} \\ \cline{3-8} 
\multicolumn{2}{|c|}{} & Offline & \multicolumn{1}{c|}{Online} & Offline & \multicolumn{1}{c|}{Online} & Offline & Online \\ \hline
\multicolumn{1}{|c|}{\multirow{4}{*}{Method}} & POD & 0.1226 & \multicolumn{1}{c|}{0.02675} & 0.8504 & \multicolumn{1}{c|}{0.02836} & 15.22 & 0.07285 \\
\multicolumn{1}{|c|}{} & GCNN & 187.7 & \multicolumn{1}{c|}{0.01383} & 385.9 & \multicolumn{1}{c|}{0.04159} & 3243 & 0.1448 \\
\multicolumn{1}{|c|}{} & CNN & 530.5 & \multicolumn{1}{c|}{0.1625} & 1830 & \multicolumn{1}{c|}{1.171} & 4494 & 3.491 \\
\multicolumn{1}{|c|}{} & FCNN & 147.2 & \multicolumn{1}{c|}{0.002186} & 322.5 & \multicolumn{1}{c|}{0.01053} & 1339 & 0.06497 \\ \hline
\end{tabular}
\bigskip
\caption{\rev{Offline and online costs (in seconds) of evaluating the ROMs corresponding to the 1-D parameterized Burgers' example with $n=10$.}  }
\label{tab:timings}
\end{table}

\subsection{Parameterized 2-D Heat Equation}

\begin{figure}
    \centering
    \includegraphics[width=0.8\linewidth]{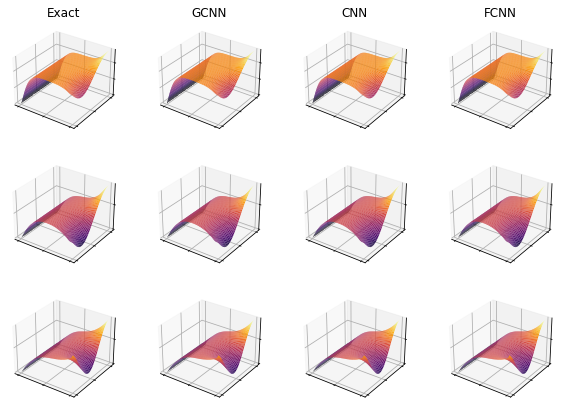}
    \caption{Example solutions to the 2-D parameterized heat equation contained in the validation set and their predictions given by the neural networks corresponding to $n=10$.  The parameters $(t,\mmu)$ are: $\left(1, 0.2, \frac{5\pi}{8}\right)$ (top), $\left(2, 0.4, \frac{5\pi}{8}\right)$ (mid), $\left(3, 0.4, \frac{7\pi}{8}\right)$ (bottom).}
    \label{fig:heatsamples}
\end{figure}

The next experiment involves the parameterized two-dimensional heat equation on a uniform grid.  In particular, let $\Omega = [0,1] \times [0,2]$, $t\in[0,4]$, $\mmu \in [0.1, 0.5] \times \left[\frac{\pi}{2}, \pi\right]$ and consider the problem
\begin{equation}\label{eq:heat}
\begin{split}
    &u_t - \Delta u = 0, \\
    &u(0,y,t) = -0.5, \\
    &u(1,y,t) = \mu_1 \cos(\mu_2 y), \\
    &u(x,y,0) = 0, \\
\end{split}
\end{equation}
where $u = u(x,y,t,\mmu)$ is the temperature profile.  Denoting the $L^2$ inner product on $\Omega$ by $(\cdot,\cdot)$, multiplying the first equation by a test function $v\in H^1_0(\Omega)$ and integrating by parts yields the standard weak equation
\[\left(u_t,v\right) + \left(\nabla u,\nabla v\right) = 0, \qquad \mathrm{for\,all}\,v\in H_0^1(\Omega), \]
which can be discretized and solved using standard finite element techniques.  In particular, choosing a finite element basis $\{\vp_i\}$ so that the discretized solution has coefficient vector $\bb{u}$ yields the linear system
\[\bb{M}\bb{u}_t = -\bb{S}\bb{u},\]
where $\bb{M} = (m_{ij}), \bb{S} = (s_{ij})$ are the symmetric mass and stiffness matrices with entries
\begin{equation}
    m_{ij} = \left(\vp_i,\vp_j\right), \qquad s_{ij} = \left(\nabla\vp_i, \nabla\vp_j\right).
\end{equation}
The present experiment examines the ability of the neural network ROMs to predict solutions to this parameterized boundary value problem.  In the full order case, $u$ is semi-discretized using bilinear $Q_1$-elements over a $32\times 64$ grid to yield $\bb{u}(t, \mmu)$ of dimension $N=2145$, and the IFISS software package \cite{elman2007} is used to assemble and solve the boundary value problem \eqref{eq:heat} using a backward Euler method with stepsize $\Delta t = 1/100$.  The system is integrated for a time period of 4 seconds at various parameters $\mmu$.  Since no classical solution to \eqref{eq:heat} exists at $t=0$, solution snapshots are stored starting from $t=0.01$. Some solutions along this heat flow contained in the validation set are displayed in Figure~\ref{fig:heatsamples}.

The parameters $\mmu$ appearing in the boundary conditions are again chosen to lie in a lattice.  In particular, the 9 values used for training are $\{ 0.1 + 0.2i, \frac{\pi}{2} + \frac{\pi}{4}j)\}$ for $0\leq i,j \leq 2$ and the 4 validation values are the lattice midpoints $\{ (0.2 + 0.2i, \frac{3\pi}{8} + \frac{\pi}{2}j)\}$.  A sample from the validation set along with its ROM reconstructions for the case $n=10$ is shown in Figure~\ref{fig:heatsamples}.  The precise architectures used are displayed in Tables~\ref{tab:GCNN}, \ref{tab:heatFCNN}, and \ref{tab:heatCNN}.  The initial learning rates are $2.5\times 10^{-3}$ for the GCNN case, $5\times 10^{-4}$ for the CNN case, and $5\times 10^{-4}$ for the FCNN case.  The early stopping criterion for network training is 100 epochs, and the mini-batch size is $n_b = 32$.

\begin{table}[]
\begin{tabular}{|c|c|c|c|}
\hline
\textbf{layer} & \textbf{input size} & \textbf{output size} & \textbf{activation} \\ \hline
\multicolumn{4}{|c|}{Samples of size (2145,1) are flattened to size (2145).} \\ \hline
1-FC & 2145 & 215 & ELU \\
1-BN & 215 & 215 & Identity \\
2-FC & 215 & $n$ & ELU \\ \hline
\multicolumn{4}{|c|}{End of encoding layers.  Beginning of decoding layers.} \\ \hline
3-FC & $n$ & 215 & ELU \\
2-BN & 215 & 215 & Identity \\
4-FC & 215 & 2145 & ELU \\ \hline
\multicolumn{4}{|c|}{Samples of size (2145) are reshaped to size (2145,1)} \\ \hline
\multicolumn{4}{|c|}{End of decoding layers.  Prediction layers listed below.} \\ \hline
5-FC & $n_\mu+1$ & 50 & ELU \\
6-FC & 50 & 50 & ELU \\
7-FC & 50 & 50 & ELU \\
8-FC & 50 & $n$ & ELU \\ \hline
\end{tabular}
\bigskip
\caption{FCNN autoencoder architecture for the 2-D heat example.}
\label{tab:heatFCNN}
\end{table}

\begin{table}[]
\begin{tabular}{|c|c|c|c|c|c|c|}
\hline
\textbf{layer} & \textbf{input size} & \textbf{kernel size} & \textbf{stride} & \textbf{padding} & \textbf{output size} & \textbf{activation} \\ \hline
\multicolumn{7}{|c|}{Samples of size (2145, 1) are zero padded to size (4096,1) and reshaped to size (1, 64, 64).} \\ \hline
1-C & (1, 64, 64) & 5x5 & 2 & SAME & (4, 32, 32) & ELU \\
2-C & (4, 32, 32) & 5x5 & 2 & SAME & (16, 16, 16) & ELU \\
3-C & (16, 16, 16) & 5x5 & 2 & SAME & (64, 8, 8) & ELU \\
4-C & (64, 8, 8) & 5x5 & 2 & SAME & (256, 4, 4) & ELU \\ \hline
\multicolumn{7}{|c|}{Samples of size (256, 4, 4) are flattened to size (4096).} \\ \hline
1-FC & 4096 & \multicolumn{3}{c|}{} & $n$ & ELU \\ \hline
\multicolumn{7}{|c|}{End of encoding layers.  Beginning of decoding layers.} \\ \hline
2-FC & $n$ & \multicolumn{3}{c|}{} & 4096 & ELU \\ \hline
\multicolumn{7}{|c|}{Samples of size (4096) are reshaped to size (256, 4, 4).} \\ \hline
1-TC & (256, 4, 4) & 5x5 & 2 & SAME & (64, 8, 8) & ELU \\
2-TC & (64, 8, 8) & 5x5 & 2 & SAME & (16, 16, 16) & ELU \\
3-TC & (16, 16, 16) & 5x5 & 2 & SAME & (4, 32, 32) & ELU \\
4-TC & (4, 32, 32) & 5x5 & 2 & SAME & (1, 64, 64) & ELU \\ \hline
\multicolumn{7}{|c|}{Samples of size (1, 64, 64) are reshaped to size (4096, 1) and truncated to size (2145, 1).} \\ \hline
\end{tabular}
\bigskip
\caption{CNN autoencoder architecture for the 2-D heat example.  Prediction layers (not pictured) are the same as in the FCNN case (c.f. Table~\ref{tab:heatFCNN}). }
\label{tab:heatCNN}
\end{table}

\begin{table}[]
\begin{tabular}{|c|c|c|c|c|l|c|c|c|c|l|}
\hline
 & \multicolumn{5}{c|}{Encoder/Decoder $+$ Prediction} & \multicolumn{5}{c|}{Encoder/Decoder only} \\ \hline
Network & $n$ & $R\ell_1$\% & $R\ell_2$\% & Size (MB) & \begin{tabular}[c]{@{}l@{}}Time per\\ Epoch (s)\end{tabular} & $n$ & $R\ell_1$\% & $R\ell_2$\% & Size (MB) & \begin{tabular}[c]{@{}l@{}}Time per\\ Epoch (s)\end{tabular} \\ \hline
GCNN & \multirow{3}{*}{3} & 7.19 & 9.21 & 0.132 & 9.5 & \multirow{3}{*}{3} & 6.96 & 9.21 & 0.0659 & 9.2 \\
CNN &  & 3.26 & 4.58 & 3.64 & 18 &  & 3.36 & 3.81 & 3.62 & 18 \\
FCNN &  & 4.75 & 6.19 & 3.74 & 3.3 &  & 4.22 & 5.69 & 3.72 & 3.1 \\ \hline
GCNN & \multirow{3}{*}{10} & 2.87 & 3.82 & 0.253 & 9.6 & \multirow{3}{*}{10} & 2.06 & 2.85 & 0.186 & 9.4 \\
CNN &  & 3.07 & 4.38 & 3.87 & 18 &  & 2.45 & 2.90 & 3.85 & 18 \\
FCNN &  & 2.96 & 3.97 & 3.76 & 3.3 &  & 2.32 & 2.92 & 3.73 & 3.1 \\ \hline
GCNN & \multirow{3}{*}{32} & 2.55 & 3.48 & 0.636 & 9.6 & \multirow{3}{*}{32} & 1.05 & 1.91 & 0.564 & 9.2 \\
CNN &  & 2.30 & 3.73 & 4.60 & 19 &  & 2.34 & 2.91 & 4.57 & 18 \\
FCNN &  & 2.65 & 4.25 & 3.80 & 3.2 &  & 1.61 & 2.31 & 3.77 & 3.2 \\ \hline
\end{tabular}
\bigskip
\caption{Numerical results for the 2-D parameterized heat equation corresponding to the prediction problem (left) and the compression problem (right).}
\label{tab:heat}
\end{table}

\begin{figure}
    \centering
    \begin{minipage}{0.5\textwidth}
        \includegraphics[width=\linewidth]{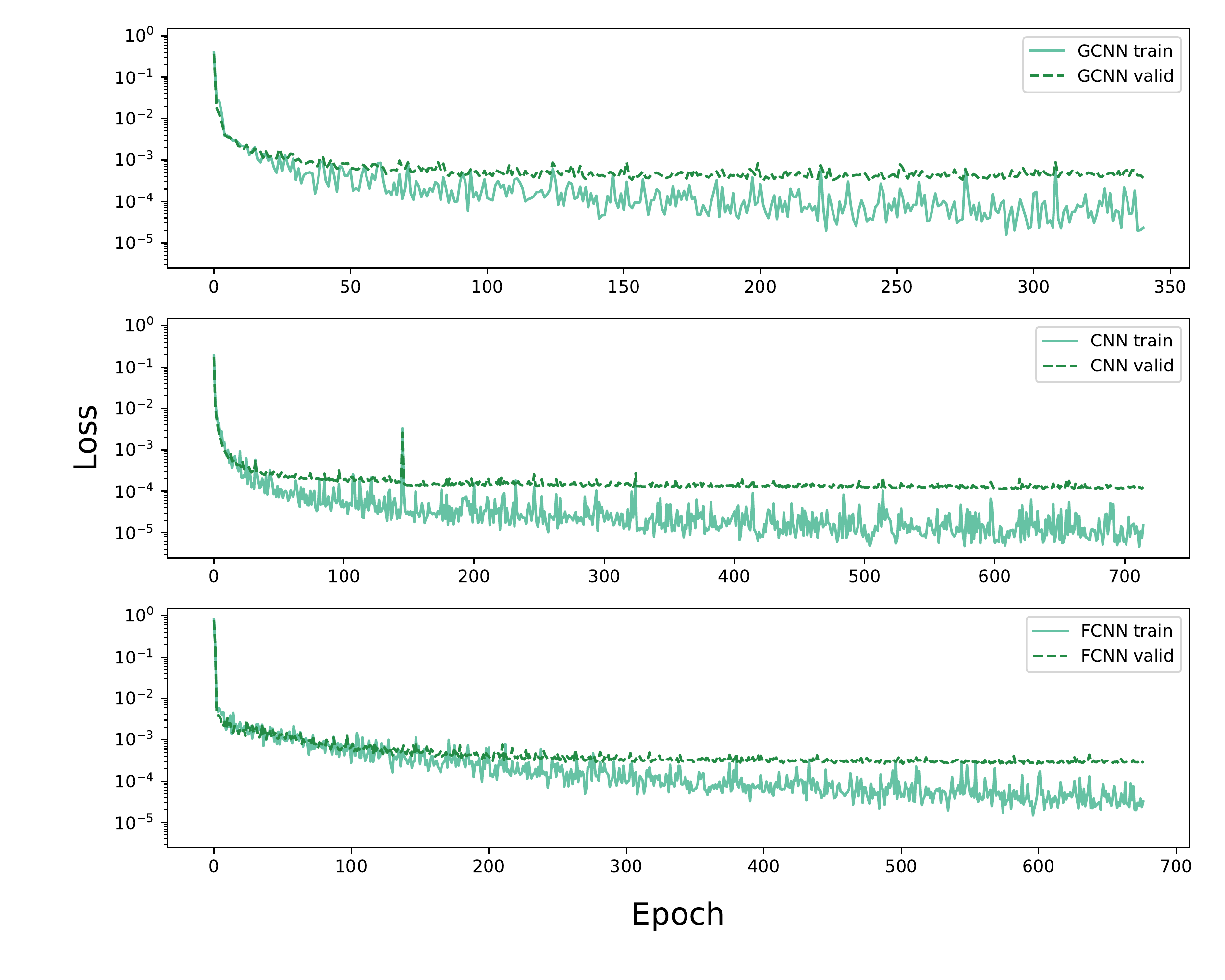}
    \end{minipage}%
    \begin{minipage}{0.5\textwidth}
        \includegraphics[width=\linewidth]{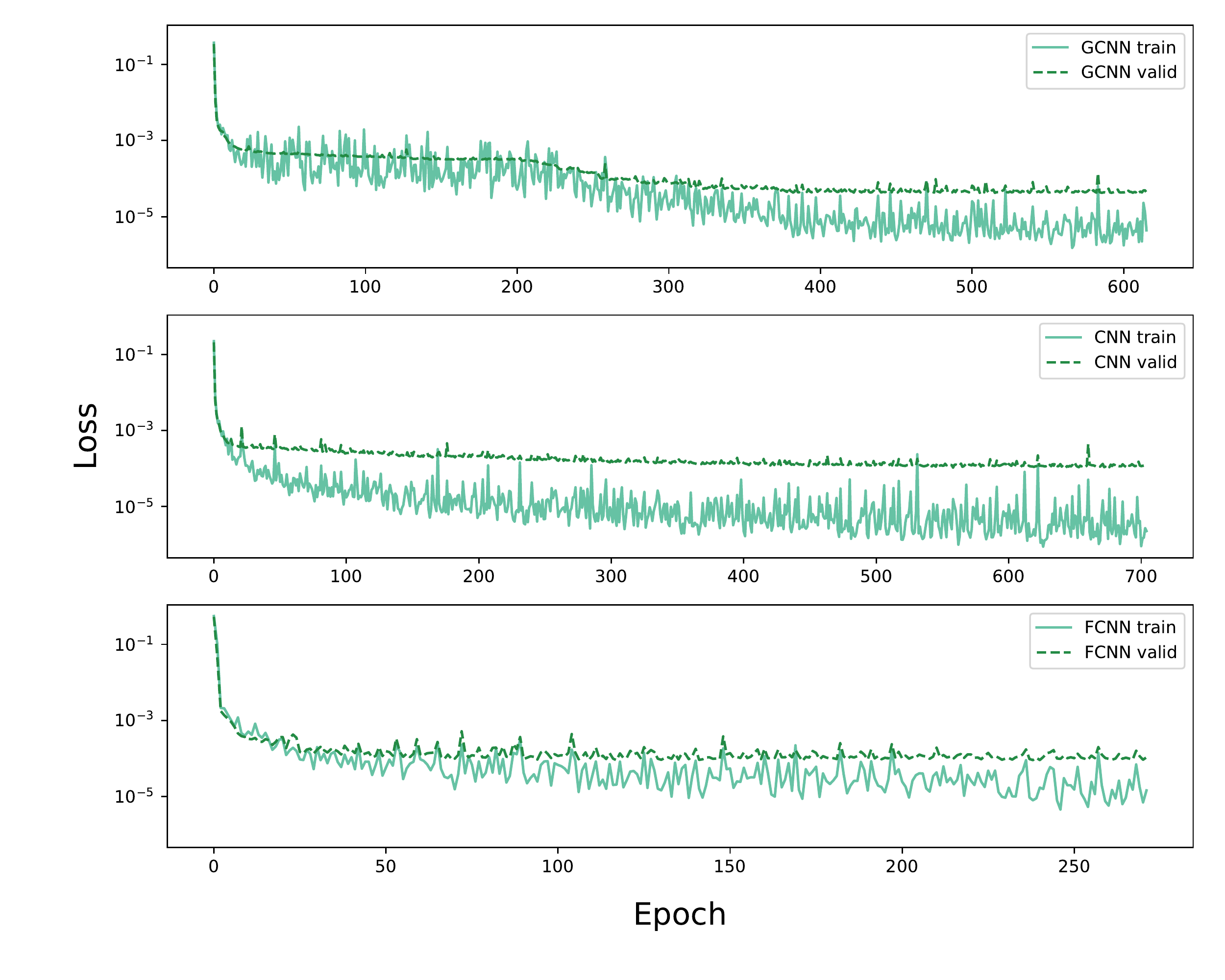}
    \end{minipage}
    \caption{Training and validation losses incurred during neural network training for the heat example.  Left: prediction problem with $n=32$.  Right: compression problem with $n=32$.}
    \label{fig:heatLoss}
\end{figure}

\begin{figure}
    \centering
    \begin{minipage}{0.5\textwidth}
        \includegraphics[width=\linewidth]{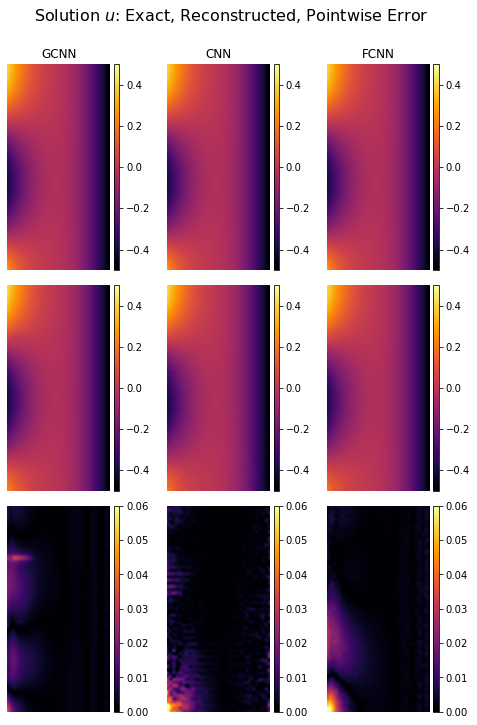}
    \end{minipage}%
    \begin{minipage}{0.5\textwidth}
        \includegraphics[width=\linewidth]{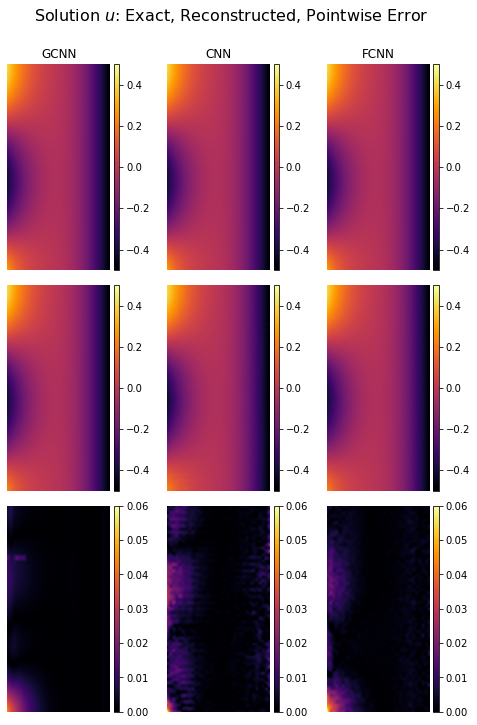}
    \end{minipage}
    \caption{Solutions corresponding to the parameter $(t,\mmu)=\begin{pmatrix}3 & 0.4 & \frac{7\pi}{8}\end{pmatrix}^\top$ on the prediction problem with $n=10$ (left) and on the compression problem with $n=32$ (right)}.
    \label{fig:heatmaps}
\end{figure}

The results for this experiment are displayed in Table~\ref{tab:heat}, which clearly illustrates the drawbacks of using the standard CNN architecture along with solution reshaping.  Not only is the computation time increased due to the introduction of almost 2000 fake nodes, but the accuracy of the CNN-ROM is significantly lower than the FCNN-ROM despite its comparable or larger memory cost.  On the other hand, the architecture based on GCNN is both the least memory consumptive and the most accurate on each problem when $n\geq 10$.  Again, results show that the GCNN-ROM is most effective on the compression problem, demonstrating significantly improved accuracy when the latent dimension is not too small.  Conversely, if a latent dimension equal to the theoretical minimum is desired, then the ROM based on deep CNN is still superior, potentially due to its multiple levels of filtration with relatively large kernels.  An example ROM reconstruction along with pointwise error plots is displayed in Figure~\ref{fig:heatmaps}.  Interestingly, the loss plots in Figure~\ref{fig:heatLoss} show that all autoencoder networks exhibit significant overfitting on this dataset, suggesting that none of the models has sufficient generalizability to produce optimal results on new data.


\subsection{Unsteady 2-D Navier-Stokes Equations}
\begin{figure}
    \centering
    \includegraphics{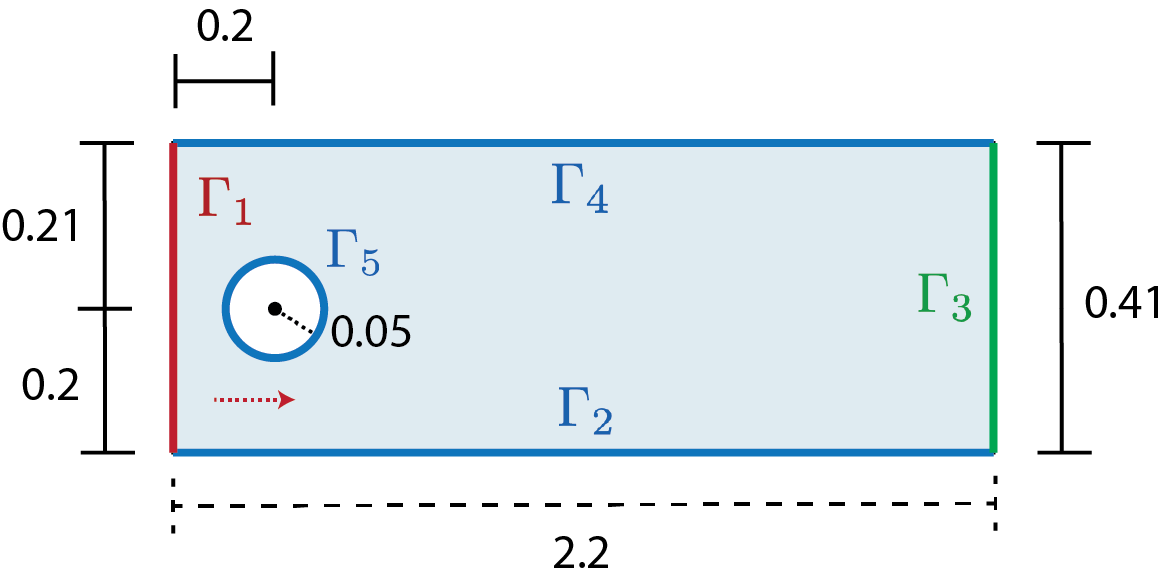}
    \caption{The domain $\Omega$ used in the Navier-Stokes example.}
    \label{fig:domain}
\end{figure}

The final experiment considers the unsteady Navier-Stokes equations on a rectangular domain $\Omega$ with a slightly off-center cylindrical hole, which is a common benchmark model for computational fluid dynamics (c.f \cite{schafer1996}).  The domain $\Omega$ is pictured in Figure~\ref{fig:domain}, and the governing system is
\begin{equation*}
    \begin{split}
        &\bb{u}_t - \nu\Delta\bb{u} + \left(\bb{u}\cdot\nabla\right)\bb{u} + \nabla p = 0, \\
        &\nabla\cdot\bb{u} = 0, \\
        &\bb{u}|_{t=0} = 0,
    \end{split}
\end{equation*}
subject to the boundary conditions
\begin{equation*}
    \begin{split}
        \bb{u} = 0 \qquad &\mathrm{on}\,\Gamma_2,\Gamma_4,\Gamma_5, \\
        \nu(\bb{n}\cdot\nabla\bb{u}) - p\bb{n} = 0 \qquad &\mathrm{on}\,\Gamma_3, \\
        \bb{u} = \begin{pmatrix}\frac{6y(0.41-y)}{0.41^2} & 0\end{pmatrix}^\top \qquad &\mathrm{on}\,\Gamma_1,
    \end{split}
\end{equation*}
where $(\bb{u},p)$ are the velocity and pressure functions and $\nu$ is the viscosity parameter.  Recall the Reynolds number
\[\mathrm{Re} = \frac{|\bb{u}|_{mean}L}{\nu},\] 
where $|\bb{u}|_{mean}$ is the average speed along the flow and $L$ is the characteristic length of the flow configuration.  The goal of this experiment is to see how well the ROM-autoencoder architectures can compress and predict solutions to these equations for a given parameter pair $(t,\mathrm{Re})$ \rev{when compared to POD and to each other.  As the meshed domain is highly irregular, this represents a good test case for the GCNN-ROM introduced before.}

To generate the training and validation data, three values of $\nu$ are selected corresponding to $(5+i)\times 10^{-4}$ for $0
\leq i \leq 2$.  This yields solutions with corresponding $\mathrm{Re} \in \{214, 180, 155\}$.  The flow is simulated starting from rest over the time interval $[0,1]$, using Netgen/NGSolve with $P_3/P_2$ Taylor-Hood finite elements for spatial discretization and implicit-explicit Euler with $\Delta t= 5\times 10^{-5}$ for time stepping. Solution snapshots are stored once every $50$ time steps, yielding a total of $400$ snapshots per example.  Note that the domain $\Omega$ is represented by an unstructured irregular mesh whose element density is higher near the cylindrical hole, yielding two solution features (the $x,y$ components of velocity) with dimension $N=10104$ equal to the number of nodes.  For the present experiments, the data corresponding to $\mathrm{Re} = 155,214$ are used to train the ROMs and the data correseponding to $\mathrm{Re}=180$ are used for validation.  The network architectures used in this case are given in Tables~\ref{tab:GCNN}, \ref{tab:nseFCNN}, and \ref{tab:nseCNN}, respectively. \rev{The POD-ROM used is detailed in Appendix~\ref{app:POD}.}  The early stopping criterion is 50 epochs, and the initial learning rate used to train each network is $5\times 10^{-4}$.

\begin{table}[]
\begin{tabular}{|c|c|c|c|}
\hline
\textbf{layer} & \textbf{input size} & \textbf{output size} & \textbf{activation} \\ \hline
\multicolumn{4}{|c|}{Samples of size (10104, 2) are flattened to size (20208).} \\ \hline
1-FC & 20208 & 2021 & ELU \\
1-BN & 2021 & 2021 & Identity \\
2-FC & 2021 & 202 & ELU \\
2-BN & 202 & 202 & Identity \\
3-FC & 202 & $n$ & ELU \\ \hline
\multicolumn{4}{|c|}{End of encoding layers.  Beginning of decoding layers.} \\ \hline
4-FC & $n$ & 202 & ELU \\
3-BN & 202 & 202 & Identity \\
5-FC & 202 & 2021 & ELU \\
4-BN & 2021 & 2021 & Identity \\
6-FC & 2021 & 20208 & ELU \\ \hline
\multicolumn{4}{|c|}{Samples of size (20208) are reshaped to size (10104, 2)} \\ \hline
\multicolumn{4}{|c|}{End of decoding layers.  Prediction layers listed below.} \\ \hline
7-FC & $n_\mu+1$ & 50 & ELU \\
8-FC & 50 & 50 & ELU \\
9-FC & 50 & 50 & ELU \\
10-FC & 50 & $n$ & ELU \\ \hline
\end{tabular}
\bigskip
\caption{FCNN autoencoder architecture for the Navier-Stokes example.}
\label{tab:nseFCNN}
\end{table}

\begin{table}[]
\begin{tabular}{|c|c|c|c|c|c|c|}
\hline
\textbf{layer} & \textbf{input size} & \textbf{kernel size} & \textbf{stride} & \textbf{padding} & \textbf{output size} & \textbf{activation} \\ \hline
\multicolumn{7}{|c|}{Samples of size (10104, 2) are zero padded to size (16384,2) and reshaped to size (2, 128, 128).} \\ \hline
1-C & (2, 128, 128) & 5x5 & 2 & SAME & (8, 64, 64) & ELU \\
2-C & (8, 64, 64) & 5x5 & 2 & SAME & (32, 32, 32) & ELU \\
3-C & (32, 32, 32) & 5x5 & 2 & SAME & (128, 16, 16) & ELU \\
4-C & (128, 16, 16) & 5x5 & 2 & SAME & (512, 8, 8) & ELU \\
5-C & (512, 8, 8) & 5x5 & 2 & SAME & (2048, 4, 4) & ELU \\ \hline
\multicolumn{7}{|c|}{Samples of size (2048, 4, 4) are flattened to size (2*16384).} \\ \hline
1-FC & 32768 & \multicolumn{3}{c|}{} & $n$ & ELU \\ \hline
\multicolumn{7}{|c|}{End of encoding layers.  Beginning of decoding layers.} \\ \hline
2-FC & $n$ & \multicolumn{3}{c|}{} & 32768 & ELU \\ \hline
\multicolumn{7}{|c|}{Samples of size (2*16384) are reshaped to size (2048, 4, 4).} \\ \hline
1-TC & (2048, 4, 4) & 5x5 & 2 & SAME & (512, 8, 8) & ELU \\
2-TC & (512, 8, 8) & 5x5 & 2 & SAME & (128, 16, 16) & ELU \\
3-TC & (128, 16, 16) & 5x5 & 2 & SAME & (32, 32, 32) & ELU \\
4-TC & (32, 32, 32) & 5x5 & 2 & SAME & (8, 64, 64) & ELU \\
5-TC & (8, 64, 64) & 5x5 & 2 & SAME & (2, 128, 128) & ELU \\ \hline
\multicolumn{7}{|c|}{Samples of size (2, 128, 128) are reshaped to size (16384, 2) and truncated to size (10104, 2).} \\ \hline
\end{tabular}
\bigskip
\caption{CNN autoencoder architecture for the Navier-Stokes example. Prediction layers (not pictured) are the same as in the FCNN case (c.f. Table~\ref{tab:nseFCNN}). }
\label{tab:nseCNN}
\end{table}

\begin{table}[]
\begin{tabular}{|c|c|c|c|c|l|c|c|c|c|l|}
\hline
 & \multicolumn{5}{c|}{Encoder/Decoder $+$ Prediction} & \multicolumn{5}{c|}{Encoder/Decoder only} \\ \hline
Network & $n$ & $R\ell_1$\% & $R\ell_2$\% & Size (MB) & \begin{tabular}[c]{@{}l@{}}Time per\\ Epoch (s)\end{tabular} & $n$ & $R\ell_1$\% & $R\ell_2$\% & Size (MB) & \begin{tabular}[c]{@{}l@{}}Time per\\ Epoch (s)\end{tabular} \\ \hline
POD & \multirow{4}{*}{2} & 10.0 & 15.85 & 0.323 & N/A & \multirow{4}{*}{2} & 6.75 & 10.0 & 0.323 & N/A \\
GCN &  & 9.07 & 14.6 & 0.476 & 33 &  & 7.67 & 12.2 & 0.410 & 32 \\
CNN &  & 7.12 & 11.1 & 224 & 210 &  & 11.2 & 17.6 & 224 & 190 \\
FCNN &  & 1.62 & 2.87 & 330 & 38 &  & 1.62 & 2.70 & 330 & 38 \\ \hline
POD & \multirow{4}{*}{32} & 2.70 & 3.97 & 5.17 & N/A & \multirow{4}{*}{32} & 0.428 & 0.674 & 5.17 & N/A  \\
GCN &  & 2.97 & 5.14 & 5.33 & 32 &  & 0.825 & 1.49 & 5.26 & 32 \\
CNN &  & 4.57 & 7.09 & 232 & 230 &  & 4.61 & 7.24 & 232 & 220 \\
FCNN &  & 1.39 & 2.64 & 330 & 38 &  & 0.680 & 1.12 & 330 & 38 \\ \hline
POD & \multirow{4}{*}{64} & 2.80 & 4.36 & 10.3 & N/A & \multirow{4}{*}{64} & 0.191 & 0.318 & 10.3 & N/A \\
GCN &  & 2.88 & 4.96 & 10.5 & 33 &  & 0.450 & 0.791 & 10.4 & 33 \\
CNN &  & 3.42 & 5.33 & 241 & 270 &  & 2.42 & 3.57 & 241 & 260 \\
FCNN &  & 1.45 & 2.64 & 330 & 38 &  & 0.704 & 1.19 & 330 & 37 \\ \hline
\end{tabular}
\bigskip
\caption{\rev{Numerical results for the Navier-Stokes example corresponding to the prediction problem (left) and the compression problem (right).  Note that the relative errors for the POD-ROM are computed with respect to the inner product defined by the finite element mass matrix.}}
\label{tab:NSE}
\end{table}

The results of this experiment are given in Table~\ref{tab:NSE}, making it clear that the trick of reshaping the nodal features in order to apply CNN carries an associated accuracy cost.  Indeed, the performance of the CNN-ROM on the prediction problem is inferior to both the GCNN-ROM and FCNN-ROM (provided the latent space is not too small) despite its much higher computational cost. Moreover, on this highly irregular domain the CNN-based architecture does not even offer benefits in the low-dimensional $n=2$ case, where the architecture based on FCNN is roughly 5 times faster to train while maintaining superior accuracy. \rev{It is also apparent that the network-based ROMs behave differently than the POD-ROM in this case.  The FCNN-ROM is consistently the most accurate throughout on the prediction problem, with this difference being the most noticeable when the dimension $n$ is very small.  On the other hand, the POD-ROM approximation is very competitive when $n=32$, but degrades with larger $n$ while the network approximations continue to improve.  This is perhaps due to the fact that 99.97 of the total squared POD singular values are captured here with only 32 modes 93.55\% with 2 modes), limiting the benefit to increasing $n$ further.}



\begin{figure}
    \centering
    \begin{minipage}{0.5\textwidth}
        \includegraphics[width=\linewidth]{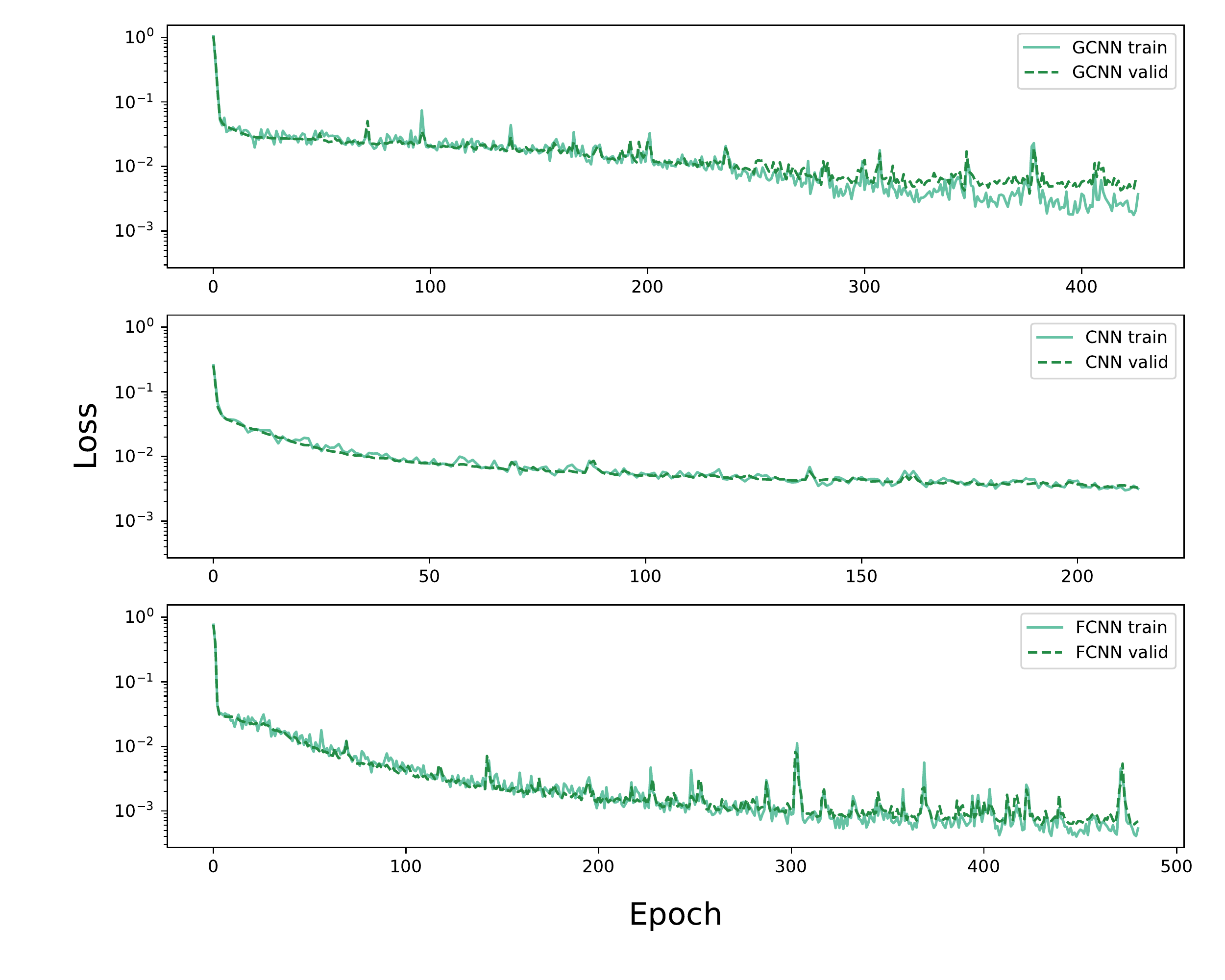}
    \end{minipage}%
    \begin{minipage}{0.5\textwidth}
        \includegraphics[width=\linewidth]{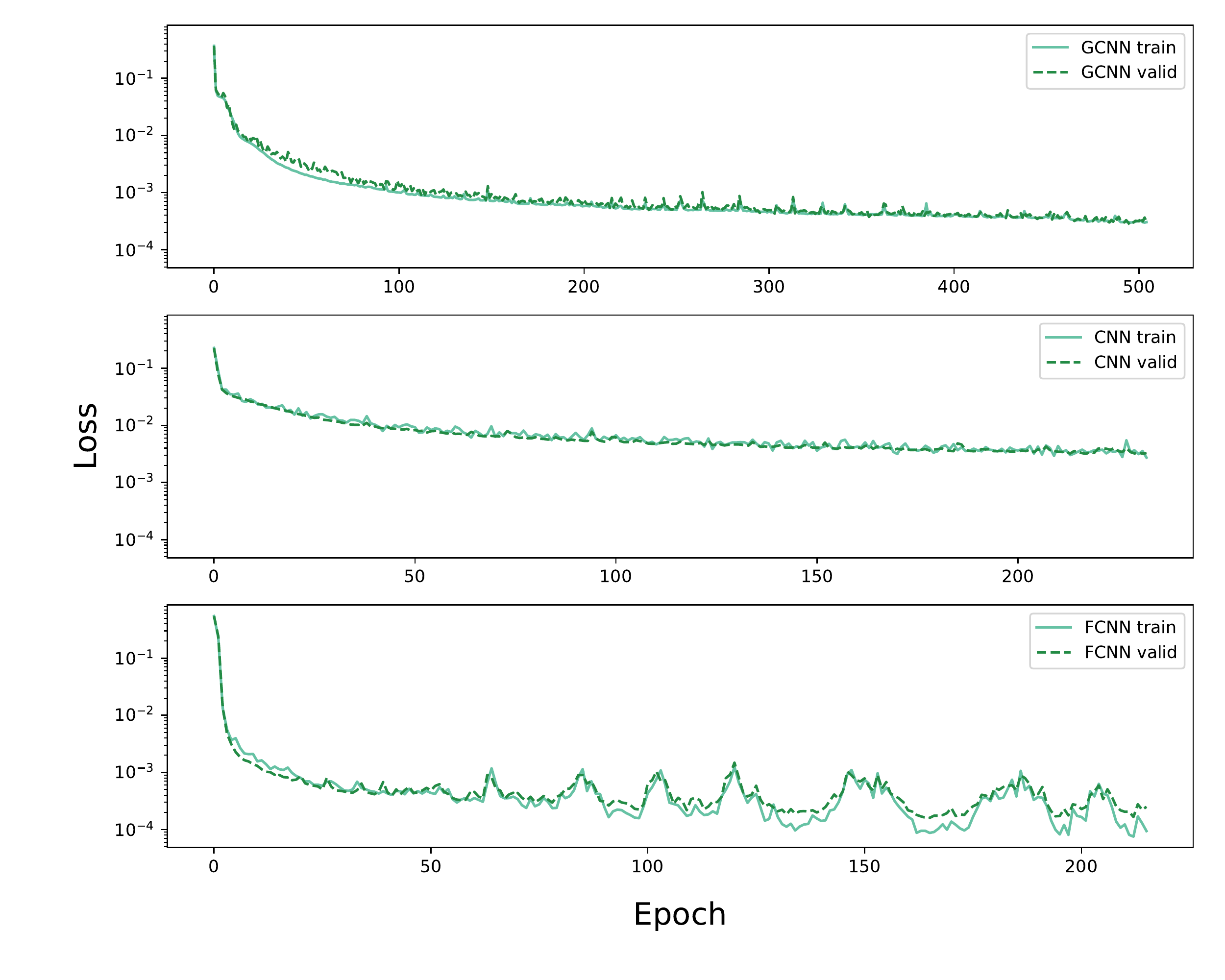}
    \end{minipage}
    \caption{Training and validation losses incurred during neural network training for the Navier-Stokes example.  Left: prediction problem with $n=32$.  Right: compression problem with $n=32$.}
    \label{fig:nseLoss}
\end{figure}
\begin{figure}
    \centering
    \begin{minipage}{\textwidth}
        \includegraphics[width=\linewidth]{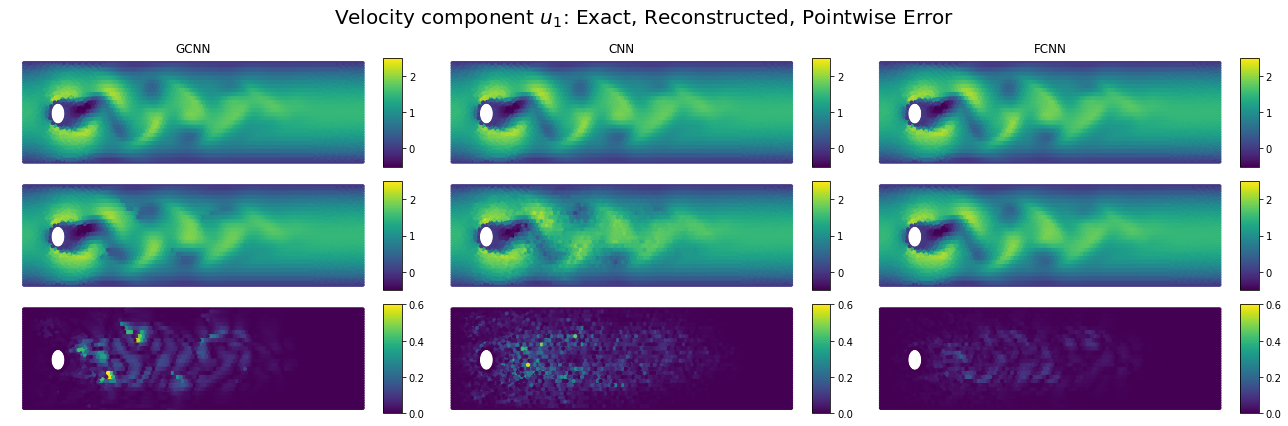}
    \end{minipage}\\
    \begin{minipage}{\textwidth}
        \includegraphics[width=\linewidth]{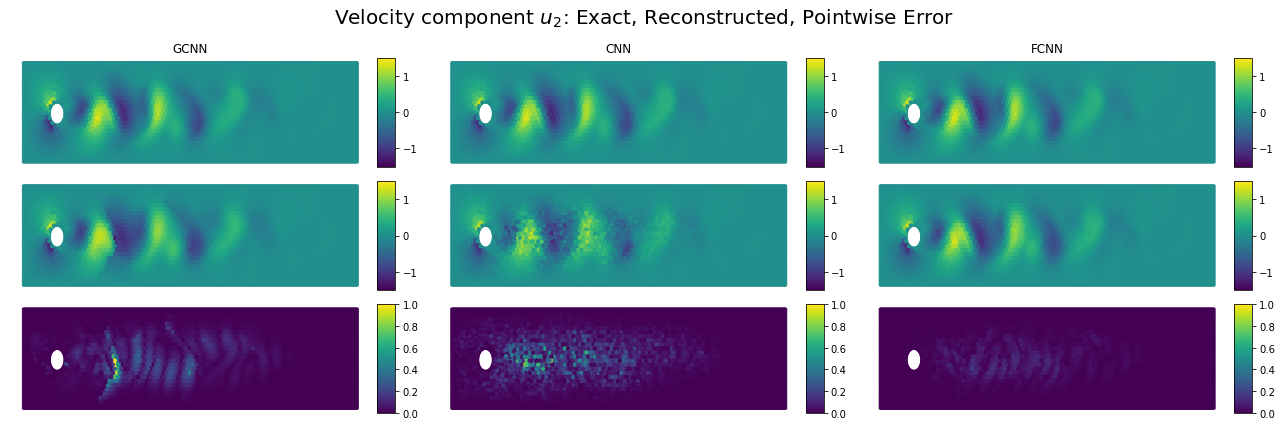}
    \end{minipage}
    \caption{Solutions to the prediction problem with $n=32$ for the parameter $(t,\mmu)=\begin{pmatrix} 0.99 & 180\end{pmatrix}^\top$.}
    \label{fig:nsepred}
\end{figure}
\begin{figure}
    \centering
    \begin{minipage}{\textwidth}
        \includegraphics[width=\linewidth]{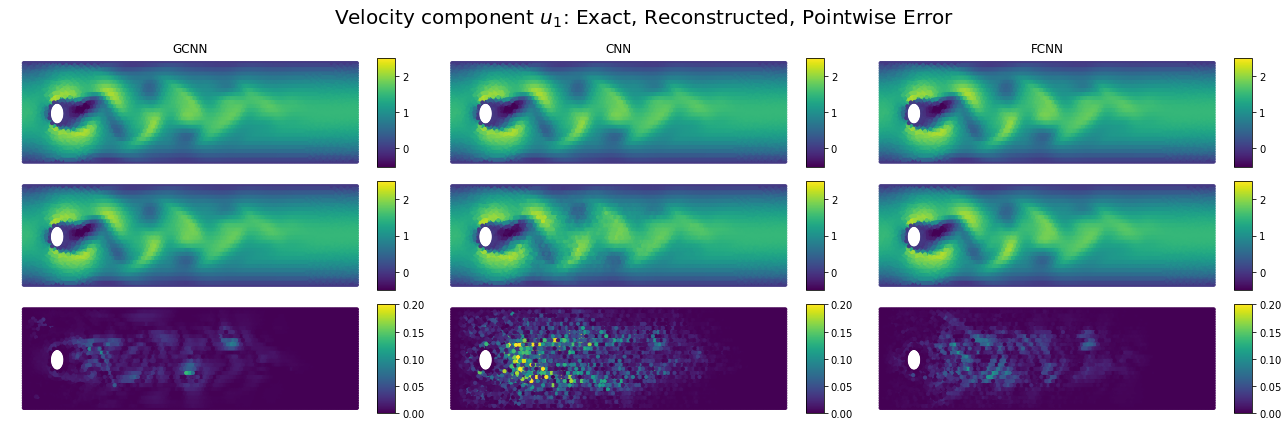}
    \end{minipage}\\
    \begin{minipage}{\textwidth}
        \includegraphics[width=\linewidth]{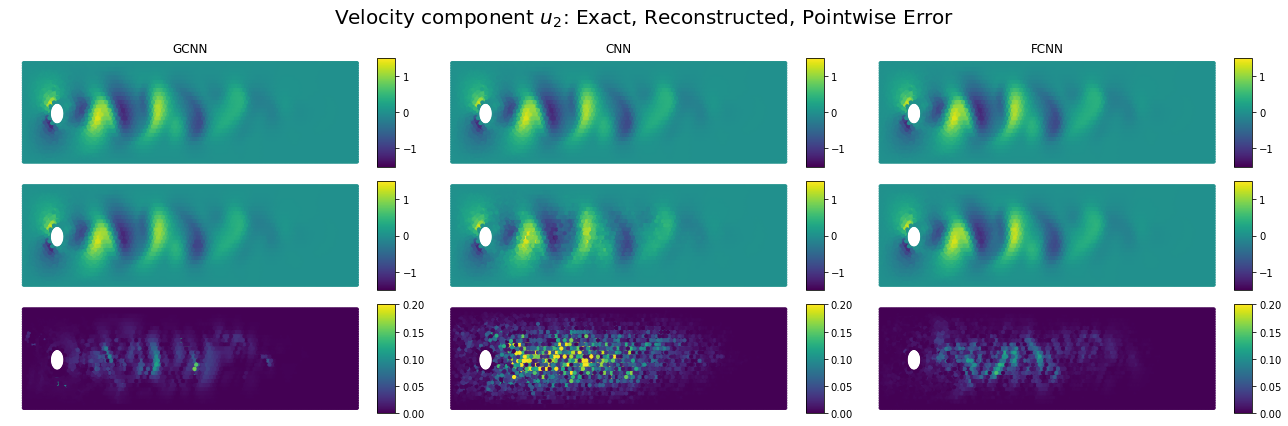}
    \end{minipage}
    \caption{Solutions to the compression problem with $n=64$ for the parameter $(t,\mmu)=\begin{pmatrix} 0.99 & 180\end{pmatrix}^\top$.}
    \label{fig:nsecomp}
\end{figure}
\begin{figure}
    \centering
    \begin{minipage}{\textwidth}
        \includegraphics[width=\linewidth]{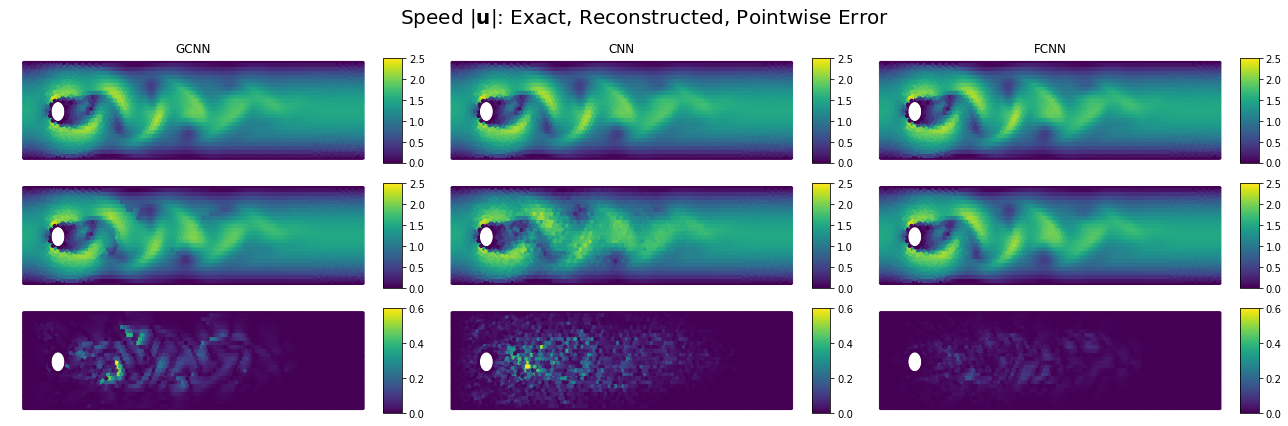}
    \end{minipage}\\
    \begin{minipage}{\textwidth}
        \includegraphics[width=\linewidth]{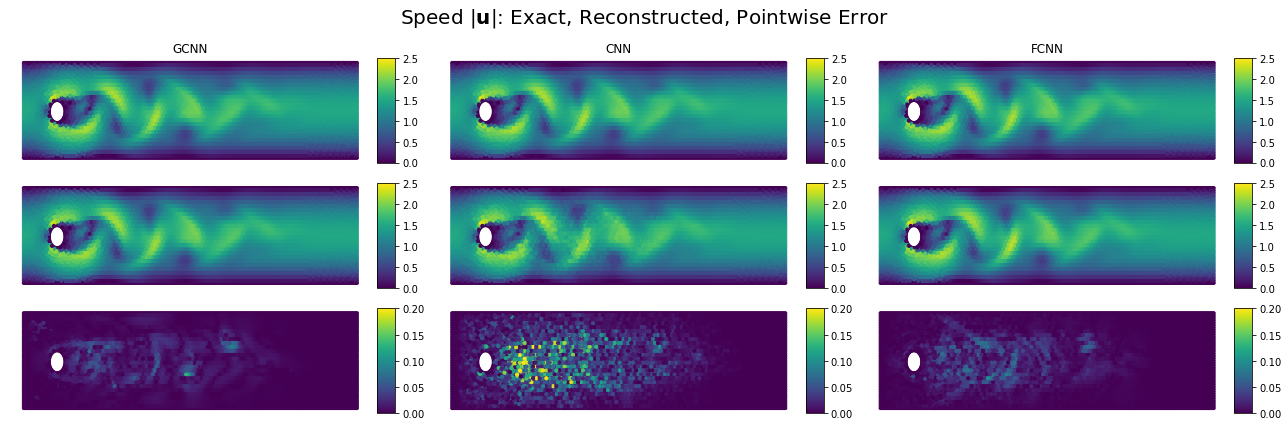}
    \end{minipage}
    \caption{Solutions to the prediction problem with $n=32$ (top) and solutions to the compression problem with $n=64$ (bottom) for the parameter $(t,\mmu)=\begin{pmatrix} 0.99 & 180\end{pmatrix}^\top$.}
    \label{fig:nsespeed}
\end{figure}
\begin{figure}
    \centering
    \begin{minipage}{\textwidth}
        \includegraphics[width=\linewidth]{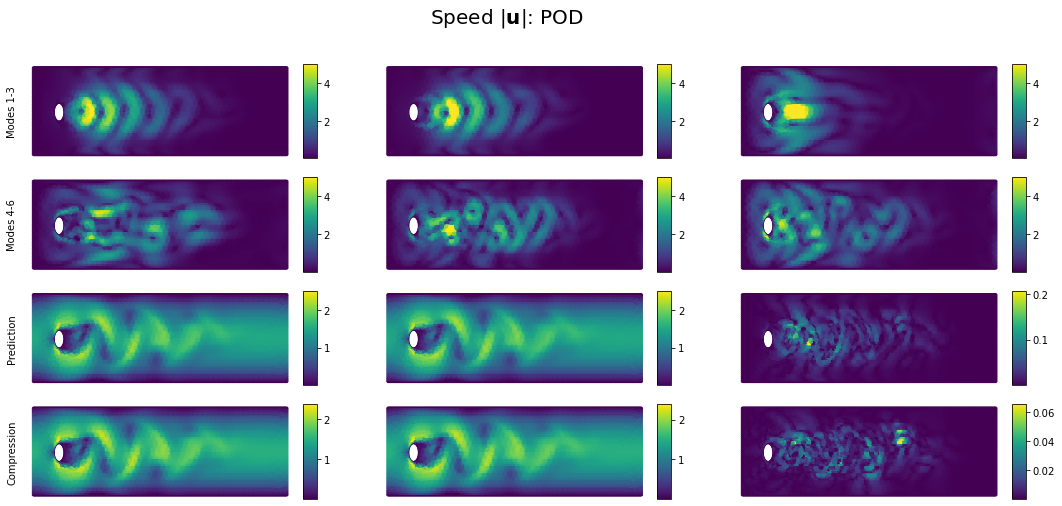}
    \end{minipage}
    \caption{\rev{Norm of the first six POD modes (rows 1 and 2) along with solutions to the prediction (row 3) and compression (row 4) problems when $(t,\mmu)=\begin{pmatrix} 0.99 & 180\end{pmatrix}^\top$.  Rows 3-4 contain the exact solution (left), the POD approximation (middle), and the pointwise error (right).}}
    \label{fig:nsepod}
\end{figure}

On the other hand, there are still interesting aspects of the network ROMs noticeable here.  As before, the GCNN-ROM struggles when $n$ is small but gains in performance as the latent dimension increases.  In fact, results indicate that the GCNN is superior to the FCNN on the compression problem when $n=64$ while also being faster to train and an order of magnitude less memory consumptive.  Although $n=64$ is far from the ``best case scenario'' of $n=2$, it is still much smaller than the original dimension of $N=10104$ and may be  useful in a real-time or many-query setting.  Interestingly, Figure~\ref{fig:nseLoss} shows that the CNN-ROM does not overfit on the prediction or compression problems, indicating that the standard CNN may not be expressive enough to handle complicated irregular data that has been aggressively reshaped.  Similarly, the GCNN-ROM does not overfit during the compression examples, although its performance scales with the dimension of the latent space, \rev{offering predictive accuracy near that of the FCNN with a memory cost near that of POD}.  Figures~\ref{fig:nsepred}, \ref{fig:nsecomp}, \ref{fig:nsespeed}, \rev{and \ref{fig:nsepod}} provide a visual comparison of the ROM reconstructions and their relative errors at $t=0.99$.   Finally, note that for this problem it may be infeasible in the case $n=2$ to generate an accurate predictive ROM using any method, since it is still an open question whether the mapping $(t, \mathrm{Re}) \mapsto \bb{u}$ is unique in this case. 

\section{Conclusion}
Neural network-based strategies for the reduced-order modeling of PDEs have been discussed, and a novel autoencoder architecture based on graph convolution has been proposed for this purpose.  Through benchmark problems, the performance of three fundamentally different autoencoder-based ROM architectures has been compared, demonstrating that each architecture has advantages and disadvantages depending on the task involved and the connectivity pattern of the input data.  \rev{In general, results indicate that the proposed graph convolutional ROM provides an accurate and lightweight alternative to fully connected ROMs on unstructured finite element data, greatly improving upon standard CNN-ROMs while maintaining a memory cost comparable to linear techniques such as POD.}  Moreover, it seems that the standard CAE is highly performing at low values of $n$ but should be avoided when the model domain is irregular, while the fully connected autoencoder exhibits strong overall performance at a higher memory cost than the alternatives.  \rev{It is also remarkable that traditional POD-based ROMs, while not strictly data-driven, can still be fast and effective when the latent dimension becomes high enough and typically exhibit more predictable dependence on dimension than the nonlinear network-based methods examined recently.}  Future work will examine how the proposed GCNN autoencoder can be combined with traditional PDE-based numerical solvers to generate solutions at unknown parameter configurations without the need for a fully connected prediction network, potentially eliminating the noted accuracy bottleneck arising from this part of the ROM.  

\section*{Acknowledgments}
This work is partially supported by U.S. Department of Energy  Scientific Discovery through Advanced Computing under grants DE-SC0020270 and
DE-SC0020418.

\bibliographystyle{ieeetr}
\bibliography{biblio.bib}

\addresseshere

\appendix

\section{Galerkin POD ROMs}\label{app:POD}
We briefly describe the Galerkin ROMs based on POD used for the examples.  Note that Einstein summation is used  throughout, so that any index which appears both up and down in a tensor expression is implicitly summed over its appropriate range.

\subsection{1-D Inviscid Burger's Equation}
To generate the POD-ROM based on the finite difference discretization in Section 4.2, the snapshot matrix $\bb{S} = \left(s^i_j\right)$ with entries  $s^i_j = w^i(t_j,\mmu_j)$ is first preprocessed by subtracting the relevant initial condition from each column, generating $\bar{\bb{S}} = \left(\bar{s}^i_j\right)$ with $\bar{s}^i_j = s^i_j - w^i(t_0,\mmu_j)$.  Factoring this through the singular value decomposition $\bar{\bb{S}} = \bb{U}\bm{\Sigma}\bb{V}^\top$, it follows that the optimal $n$-dimensional linear solution subspace (in the sense of variance maximization) is the span of the matrix $\bb{U}_n$ containing the first $n$ columns of $\bb{U}$.  The POD approximation $\tilde{\bb{w}} \approx \bb{w}$ is then expressed as \[\tilde{\bb{w}} = \bb{w}_0 + \bb{U}_n\hat{\bb{w}},\]
where $\bb{w}_0(\mmu) = \bb{w}(t_0,\mmu)$ and $\hat{\bb{w}}\in \mathbb{R}^n$ denotes the reduced-order state.  Substituting this into the (discretization of) system \eqref{eq:burgers} yields the reduced-order system
\begin{equation}\label{eq:burgersROM}
    \begin{split}
        \frac{\hat{\bb{w}}_{k+1}-\hat{\bb{w}}_k}{\Delta t} + \frac{1}{2}\left(\bb{a} + \bb{B}\hat{\bb{w}}_k + \bb{C}(\hat{\bb{w}}_k,\hat{\bb{w}}_k)\right) &=  0.02 \bb{U}_n^\top e^{\mu_2\bb{x}}, \\
        \hat{\bb{w}}_0 &= \bb{U}_n^\top(\bb{w}_0-\bb{w}_0) = \bm{0}.
    \end{split}
\end{equation}
where $0 \leq k \leq N_t-1$, $e^{\mu_2\bb{x}}$ acts component-wise, and $\bb{a},\bb{B},\bb{C}$ are vector-valued quantities which can be precomputed.  In particular, let $\bb{D}_x \in \mathbb{R}^{N\times N}$ denote the bidiagonal forward difference matrix with entries $\{0,1/\Delta x,...,1/\Delta x\}$ on its diagonal and $\{-1/\Delta x,..,-1/\Delta x\}$ on its subdiagonal. 
A straightforward computation then yields 
\begin{align*}
\bb{a} &= \bb{U}_n^\top\bb{D}_x(\bb{w}_0)^2, \\
\bb{B} &= 2\bb{U}_n^\top\bb{D}_x\mathrm{Diag}(\bb{w}_0)\bb{U}_n, \\
\bb{C} &= \bb{U}_n^\top\bb{D}_x \mathrm{diag}_{1,3}(\bb{U}_n \otimes \bb{U}_n),
\end{align*}
where $(\bb{w_0})^2$ is a component-wise operation taking place in $\mathbb{R}^N$, $\mathrm{Diag}(\bb{w}_0)$ indicates construction of the diagonal matrix built from $\bb{w}_0$, and $\mathrm{diag}_{1,3}(\bb{U}_n\otimes\bb{U}_n)$ indicates extraction of the ``diagonal'' tensor formed from $\bb{U}_n\otimes\bb{U}_n$ when components 1 and 3 are the same, i.e. the rank 3 tensor with components $u^i_j u^i_k$.  To generate the POD-ROM solution to the prediction problem at a particular $\mmu$ in the test set, the low-dimensional system \eqref{eq:burgersROM} is solved for the reduced-order state $\hat{\bb{w}}\in\mathbb{R}^{n\times N_t}$, and the high-dimensional solution is rebuilt as $\tilde{\bb{w}} = \bb{w}_0 + \bb{U}_n\hat{\bb{w}}$ (where $\bb{w}_0$ is the relevant initial condition).  Solutions to the compression problem are generated through simple projection/expansion as $\tilde{\bb{w}} = \bb{w}_0 + \bb{U}_n\bb{U}_n^\top\left(\bb{w}-\bb{w}_0\right)$.




\subsection{Unsteady 2-D Navier-Stokes}
The POD-ROM for the Navier-Stokes equations is based on the finite element discretization used to generate the FOM snapshot data.  Beginning with the continuous velocity $\bb{u}$, the Galerkin assumption in this case is that 
\[\bb{u}(x,t) \approx\tilde{\bb{u}}(x,t) =  \bar{\bb{u}}(x) + u^I(t)\ppsi_I(x)\] 
for a time invariant ``central velocity'' $\bar{\bb{u}}$ and $N$ shape functions $\ppsi_I$ localized on the discrete domain.  From this, POD seeks a collection of $n$ variance-maximizing basis functions $\pphi_a(x)$ such that $u^I(t) = \phi^I_a\hat{u}^a(t)$ and hence $u^I\ppsi_I = \hat{u}^a\phi^I_a\ppsi_I = \hat{u}^a\pphi_a$.  Discretely, it is useful to consider the velocity $\bb{u}$ as a $2N$-length vector of nodal values, where the components $u_1, u_2$ are vertically concatenated.  In this case, the mean velocity $\bar{\bb{u}}$ is also a $2N$-vector containing the mean of each component separately, and the FOM snapshot matrix is a $2N\times N_s$ array.


Using $\bb{M}=\bb{R}^\top\bb{R}$ to denote the $(2N\times 2N)$ block diagonal mass matrix and its Cholesky factorization, it can be shown that the POD basis is related to the left singular vectors of the modified snapshot matrix $\tilde{\bb{S}} = \bb{R}\bb{S} = \tilde{\bm{\Phi}}\bm{\Sigma}\bb{V}^\top$ where $\bb{S} = (s^I_\alpha) \in \mathbb{R}^{2N\times N_s}$ is the mean-centered snapshot matrix with entries $s^I_\alpha = u^I(t_\alpha, \mmu_\alpha) - \bar{u}^I(\mmu_\alpha)$.  In particular, the first $n$ columns $\bm{\Phi}_n$ of the matrix $\bm{\Phi}$ satisfying $\bb{R}\bm{\Phi} = \tilde{\bm{\Phi}}$ are $\bb{M}$-orthonormal by construction and comprise the dimension $n$ POD basis for $\bb{S}$.  This represents the best possible $n$-dimensional linear approximation to snapshot dynamics, and solutions to the compression problem are readily constructed from it as 
\[\tilde{\bb{u}} = \bar{\bb{u}} + \bm{\Phi}_n\bm{\Phi}_n^\top\bb{M}\left(\bb{u}-\bar{\bb{u}}\right),\]
where $\bb{u},\bar{\bb{u}}$ can now depend on arbitrary $(t,\mmu)$ pairs.



Using $\left(\cdot,\cdot\right)$ to denote the $L^2$ inner product on the domain $\Omega$, integrating the Navier-Stokes equations by parts and using the boundary conditions yields the familiar weak form
\begin{align*}
    \left(\bb{u}_t,\ppsi\right) + \nu\left(\nabla \bb{u},\nabla\ppsi\right) + \left(\bb{u}\cdot\nabla\bb{u},\ppsi\right) - \left(p,\nabla\cdot\ppsi\right) &= 0, \qquad \ppsi \in H^1_0(\Omega; \mathbb{R}^2), \\
    -\left(\eta,\nabla\cdot\bb{u}\right) &= 0 \qquad \eta \in L^2(\Omega),
\end{align*}
which represents the full-order system.  A reduced-order system can be constructed in a similar fashion by using the POD decomposition.  Substituting $\tilde{\bb{u}}\approx\bb{u}$ into the FOM, symmetrizing the weak Laplace term, and testing against the POD basis $\bm{\Phi}_n$ yields
\[ \left(\tilde{\bb{u}}_t, \pphi\right) +\nu\left(\nabla\tilde{\bb{u}}+\nabla\tilde{\bb{u}}^\top,\nabla\pphi\right) + \left(\tilde{\bb{u}}\cdot\nabla\tilde{\bb{u}},\pphi\right) = 0, \qquad \pphi \in \bm{\Phi}_n, \]
where $\nabla\cdot\nabla\bb{u}^\top = u^{ijk}\bb{e}_k\cdot(\bb{e}_i\otimes\bb{e}_j) = u^{ij}_{\,\,\,\,i}\bb{e}_j = \partial^j u^i_i\bb{e}_j = \nabla(\nabla\cdot\bb{u}) = 0$ since the mixed partial derivatives of $\bb{u}$ commute.  Note that  incompressibility of $\tilde{\bb{u}}$ is automatically satisfied as the mean velocity $\bar{\bb{u}}$ and POD basis $\pphi$ are linear combinations of the snapshots.  Moreover, in the ROM it is no longer necessary to keep track of the pressure $p$.  Further computation using the $(\cdot,\cdot)$-orthogonality of the POD basis yields the reduced-order ODE system 
\[ \hat{u}_{c,t} = -r_c - \hat{u}^a S_{ac} - \hat{u}^b\hat{u}^a T_{bac},\]
for the components of $\hat{\bb{u}}$, where the quantities
\begin{align*}
    r_c &= \left(\bar{\bb{u}}\cdot\nabla\bar{\bb{u}}, \pphi_c\right) + \nu\left(\nabla\bar{\bb{u}}+\nabla\bar{\bb{u}}^\top, \nabla\pphi_c\right), \\
    S_{ac} &= \left(\bar{\bb{u}}\cdot\nabla\pphi_a + \pphi_a\cdot\nabla\bar{\bb{u}}, \pphi_c\right) + \nu\left( \nabla\pphi_a + \nabla\pphi_a^\top, \nabla\pphi_c\right), \\
    T_{bac} &= \left(\pphi_b\cdot\nabla\pphi_a,\pphi_c\right),
\end{align*}
are precomputed.  To generate the POD comparisons in Section 4.4, the above system is solved using a simple forward Euler scheme with $\Delta t = 2.5\times 10^{-5}$, which is 100 times smaller than the time step between FOM snapshots.  Solutions to the prediction problem are then given simply by $\tilde{\bb{u}} = \bar{\bb{u}} + \bm{\Phi}_n\bb{\hat{u}}$.  Note that in this case the notion of  error is best captured by the norm $\bb{u} \mapsto \left(\bb{u}^\top\bb{M}\bb{u}\right)^{1/2}$ defined by the mass matrix, so all relative error metrics are reported in this norm.

\end{document}